\documentclass[ejs]{imsart}

\RequirePackage[OT1]{fontenc}
\usepackage{amsfonts}
\usepackage{amsmath,amsthm}
\usepackage{amssymb}
\usepackage{bbm,amsbsy}
\usepackage{geometry}

\doi{10.1214/154957804100000000}
\pubyear{0000}
\volume{0}
\firstpage{1}
\lastpage{1}

\usepackage{pgf,tikz,pgfplots} 
\pgfplotsset{compat=1.14}
\usepackage{mathrsfs}
\usetikzlibrary{arrows}
\startlocaldefs
\numberwithin{equation}{section}
\theoremstyle{plain}
\newtheorem{theorem}{Theorem}[section]
\newtheorem{lemma}{Lemma}[section]
\newtheorem{rem}{Remark}[section]
\newtheorem{cor}{Corollary}[section]
\newtheorem{defini}{Definition}[section]

\newcommand{\esp}{\mathbb{E}}

\newcommand{\pr}{\mathbb{P}}

\newcommand{\inovField}{ \displaystyle {\left( \varepsilon_{t} \right)}_{t \in \mathbb{Z}^\kappa}}
\newcommand{\model}{\hat{f}}
\newcommand{\remp}[1]{\mathcal{R}_{emp}(#1)}

\newcommand{\rth}[1]{R(#1)}

\newcommand{\card}[1]{\text{Card}(#1)}
\newcommand{\SI}{S_{\mathcal{I}}}
\newcommand{\SItilde}{\Tilde{S}_{\mathcal{I}}^{[d]}}
\newcommand{\SItildep}{\Tilde{S}_{\mathcal{I}}^{[d] \prime}}
\newcommand{\SItildei}{\Tilde{S_{\mathcal{I}}}^{[d,i]}}
\newcommand{\dm}[1]{\textit{diam} \left( #1 \right)}
\newcommand{\hd}[1]{\Tilde{H}^{[#1]}}
\newcommand{\Xt}[2]{\Tilde{X}_{#1}^{[#2]}}
\newcommand{\Xtt}[2]{\Tilde{X}_{#1}^{[#2] \prime}}
\endlocaldefs

\begin{document}

\begin{frontmatter}

\title{Concentration inequalities for non-causal random fields\thanksref{t1}}
\thankstext{t1}{We thank Paul Doukhan, Joseph Rynkiewicz and Xiequan Fan for their advice. 
\\
This work was partly involved in the CY Initiative of Excellence
(grant "Investissements d'Avenir" ANR-16-IDEX-0008),
Project "EcoDep" PSI-AAP2020-0000000013.}

\author{\fnms{R\'emy} 
\snm{Garnier}
\ead[label=e1]{remy.garnier@ext.cdiscount.com}}
\address{CY University - AGM UMR 8088 \\
 2 Bd. Adolphe Chauvin. 95000 Cergy-Pontoise, France }
 \address{CDiscount\\  120-126 Quai de Bacalan, 33300 Bordeaux, France \\
 \printead{e1}}
\author{
 \fnms{Rapha\"el} 
 \snm{Langhendries}
 \ead[label=e2]{r.langhendries@gmail.com}}\\
 \address{University Paris 1 Panthéon-Sorbonne - SAMM \\
 90, rue de Tolbiac 75013 PARIS CEDEX 13 - France}
 \address{Safran Aircraft Engines,\\ 77550 Moissy-Cramayel, France\\
 \printead{e2}}
\runtitle{Concentration inequalities for non-causal random fields}
\runauthor{Garnier & Langhendries}

\begin{abstract}%
    Concentration inequalities are widely used for analyzing machines learning algorithms. However, current concentration inequalities cannot be applied to some of the most popular deep neural networks, notably in natural language processing. This is mostly due to the non-causal nature of such involved data, in the sense that each data point depends on other neighbor data points. In this paper, a framework for modeling non-causal random fields is provided and a Hoeffding-type concentration inequality is obtained for this framework. The proof of this result relies on a local approximation of the non-causal random field by a function of a finite number of i.i.d. random variables. 
\end{abstract}%

\begin{keyword}
\kwd{non-causal random fields}
\kwd{learning in non-i.i.d scenarios}
\kwd{Hoeffding's inequality}
\kwd{local dependence}
\kwd{learning theory}
\end{keyword}

\tableofcontents
\end{frontmatter}

\section{Introduction}
Concentration inequalities are widely used in statistical learning theory. For example, model selection techniques rely heavily on concentration inequalities \cite{massart2007concentration}. They have also been used for high dimensional procedures \cite{alquier2020high,bickel2009simultaneous} or for studying various machine learning framework, such as time series prediction \cite{kuznetsov2015learning}, online machine learning \cite{sanchez2015time} or classification problems \cite{freund2004generalization}.
\\
Many concentration inequalities have been proposed under different scenarios and assumptions. The simpler case corresponds to the \emph{independent and identically distributed hypothesis}.  Interested readers may consult \cite{boucheron2013concentration} for an overview on concentration inequalities in this case.  

However, in many machine learning applications, the i.i.d. assumption does not hold. Moreover, the i.i.d. cannot be tested from any data analysis. It is the case, for instance for time series which exhibit inherent temporal dependence. 
A simple way to model temporal dependence for time series is to suppose that data $(X_t)$ are generated from a Markov chain, i.e., they verify:

\begin{equation}
X_t = F(X_{t-1}, \dots, X_{t-i}, \varepsilon_t),
\;
\text{for a function $F$ and i.i.d. innovations  $(\varepsilon_t)_{t \in \mathbb{Z}}$.}
\label{MarkovianMasterEqu}
\end{equation}

In this case, many approaches have been used to establish concentration inequalities. These approaches include renewal technique (see: \cite{bertail_new_2019}), Marton coupling in \cite{paulin2015concentration} and martingale decomposition (see~\cite{alquier2019exponential,ChenWU2018,dedeckerDeviation2019,dedeckerdeviation2015,kontorovichconcentration2008}).

However, Markov chains are not sufficient to model any type of dependence. Indeed some data may exhibit non-causal dependence. In the uni-dimensional case, it means that one data point does not merely depend on past data points, but also on future data.
\\
For example, this situation occurs for textual data. A key task involving textual data is the \emph{completion problem}. It consists in filling blanks in a text using surrounding words. This task is achieved creating a language model, i.e., a probability distribution of words for a given context (past and future words). As long as both past and future words are needed to learn a consistent language model, textual data are a typical example of data generated by a \emph{non causal} process.
\\
In practice, non-causal models are already used to learn language models. It includes bidirectional neural network \cite{schuster1997bidirectional}. They have recently received a lot of attention for their performances in Natural Language Processing. In particular, the BERT model \cite{devlin2018bert} has become a staple for a very large range of NLP tasks, such as translation, part-of-speech tagging, sentiment analysis. However, despite their success in practical applications, it lacks a theoretical framework to analyze such non-causal models. 

If  the dimension of the lattice of random variables  increases, we obtained a  \emph{random field}. The natural extension of Markov chains to random fields leads to causal random fields: here the dependence are propagated along preferential directions (see \cite{doukhan2017class} for an example of application). Non-causal random fields  appear more naturally in many applications. For instance, pictures are naturally generated by a non-causal random field defined over a two-dimensional lattice. In this case, the completion problem consists in filling missing pixels using neighboring pixels \cite{ballester2001filling}. It had to use information from each direction (up, down, left, right) in order to be able to complete the missing pixel. However, contrary to the case of textual data, such pixel completion techniques are not in the state-of-the-art for image processing. Another application of importance is the case of completing geographical data sets which make sense for an ecology setting (see {\tt http://doukhan.u-cergy.fr/EcoDep.html}), for which applications are of a fundamental importance.

\subsection{Our contribution}

In this article, we extend the Markovian framework presented in Eq. \eqref{MarkovianMasterEqu} to handle non-causal data. In the 1-dimensional case, we assume that the model is  solution of the following equations 

\[
X_t = F(X_{t-s}, \dots, X_{t-1}, X_{t+1}, \dots, X_{t+s}, \varepsilon_t),
\;
\text{with i.i.d. $(\varepsilon_t)_{t \in \mathbb{Z}}$.}
\]

We immediately generalize this approach to random fields indexed by $\mathbb{Z}^{\kappa}$ ,i.e., the multi-dimensional case. In this case, data are solution of the following equation 

\[
X_t = F((X_{t+s})_{s \in \mathcal{B}},\varepsilon_t),
\;
\text{for some neighborhood $\mathcal{B}$ and i.i.d. $(\varepsilon_t)_{t \in \mathbb{Z}}$.}
\]

This type of random field has been introduced by \cite{doukhan2007fixed} in conditions for the existence and uniqueness of such solution are provided. 

We aim at establishing concentration inequalities on these processes using realistic hypotheses on data. In particular, we want our hypotheses to be reasonable for textual data. Our main results are Hoeffding-type inequalities. The main hypothesis for these inequalities are relaxed versions of the contraction assumption used in \cite{alquier2019exponential,dedeckerDeviation2019,dedeckerdeviation2015}.
\\
Our method relies on a convenient local approximation of a non causal random field by a function of finitely many i.i.d. random variables. We also provide some examples of such random fields, comparing with other frameworks, such as weak dependence and mixing techniques.
Finally, we provide a simple  application of our results to model selection, inspired by \cite{Lugosi2002}.

\subsection{Outline of the paper}

In Section \ref{Sec:Model}, we introduce non-causal random fields proposed in \cite{doukhan2007fixed}. We present our hypotheses, which are a non-causal version of the hypotheses of \cite{alquier2019exponential} or  \cite{dedeckerdeviation2015}. We also provide some examples of model satisfying our framework, then we define the statistic $\SI$ of interest for which we aim at proving a concentration inequality.
Main results may be found in Section \ref{Sec:Main} as well as some immediate applications to machine learning and comparisons with other concentration inequalities. 

The rest of the paper is dedicated to the proof of main results. 
\\
In the section \ref{Sec:Concentration}, we introduce an approximation $\SItilde$ of $\SI$ which only depends on a \emph{finite} number of independent variables. We also establish a result evaluating the quality of this approximation. 
\\
Then, in section \ref{Sec:Deviation} we prove the main concentration inequality using the McDiarmid's inequality proposed by \cite{combesExtension2015} on $\SItilde$. 


\section{Model}
\label{Sec:Model}
In this section, we will present a model inspired by \cite{doukhan2007fixed} in order to consider non-causal random field. Then, we introduce our hypotheses and the targeted statistic $\SI$.

\subsection{Some definitions and notations}

\label{1notations}

From now, all random variables are defined on a probability space $(\Omega, \mathcal{F}, \mathbb{P})$.

\paragraph{Dimension of the random field.}
Let $\kappa \in \mathbb{N}$ denote the \emph{dimension} of the random field of interest. 
This \emph{dimension}  should not be confused the classical dimension that appears in high-dimensional statistics, i.e., the number of parameters of the model. Here, the number of parameters increases exponentially with the dimension $\kappa$.

\paragraph{Probabilistic setting.}
 Let $\mathcal{X}$ be a Banach space endowed with a norm $\|\cdot\|$. We define the $m-$norm $\|X\|_m$ of a random variable $X$ as
\[
    \| X \|_m = (\esp\|X\|^m)^{\frac{1}{m}} \quad \text{with $\|\cdot\| $ is a norm on $\mathcal{X}$}.
\]
We also use the uniform norm $\displaystyle \|X\|_{\infty} = \sup \|X\| = \lim_{m \to \infty} \|X\|_m$.

\paragraph{Neighborhoods $\mathcal{V}(\delta,s)$ and $\mathcal{B}$.}
 Let $\delta$ be a $\kappa$-tuple of non-negative integer, i.e., $\delta = (\delta_1, \dots, \delta_{\kappa}) \in \mathbb{N}^{\kappa}$ and $s$ be a $\kappa$-tuple of integers, i.e., $s = (s_1,\dots s_{\kappa}) \in \mathbb{Z}^{\kappa}$. The $\delta$-neighborhood of $s$, $\mathcal{V}(\delta,s)$ is defined as the following $\kappa$-orthotope
\[
\mathcal{V}(\delta,s) = \{ t = (t_1, \dots, t_\kappa) \in (\mathbb{Z})^{\kappa} / \, \forall i, s_i - \delta_i \leq t_i \leq s_i + \delta_i \}.
\]

Thereafter, we will study a specific neighborhood  $\mathcal{B} = \mathcal{V}(\delta,0) \backslash \{ 0 \} $ for a fixed value of $\delta$.
\\
Moreover we will denote by $n_B$ the cardinal of $\mathcal{V}(\delta,0) = \mathcal{B} \bigcup \{0\}$ (thus $n_B = card(\mathcal{B}) + 1$).

\paragraph{Innovations} $\pmb{\varepsilon} = \inovField$.
Let $\inovField$ be a random field indexed by $\mathbb{Z}^\kappa$ of \emph{independent and identically distributed} random variables $\varepsilon_t$ on a Banach space $E$. To shorten the notation, we denote by $\pmb{\varepsilon} = \inovField$ the whole random field. \\
Additionally, we define $\mu_\varepsilon$ the probability distribution of one random variable $\varepsilon_t$ and $\mu = \underset{ t \in \mathbb{Z}^\kappa}{\bigotimes} \mu_\varepsilon $, the product of these distributions on $E^{\mathbb{Z}^\kappa}$. Then, $\mu$ is the distribution of $\pmb{\varepsilon}=\inovField$.

\subsection{Local non-causal relations}

In this paper, we will study the $\kappa$-dimensional non-causal random field  $(X_t)_{ t \in \mathbb{Z}^\kappa}$. This random field presents a \emph{local dependence}, for each $t \in \mathbb{Z}^\kappa$, $X_t$ depends on its neighborhood indexed by $\mathcal{V}(\delta,t)$ and on the innovation $\varepsilon_t$. \\
Formally, we assume it exists a function $F :  \mathcal{X}^{\mathcal{B}} \times E \mapsto \mathcal{X}$ such that $(X_t)_{ t \in \mathbb{Z}^\kappa}$ is a stationary solution of the following equation

\begin{equation}
\forall t \in \mathbb{Z}^\kappa, X_t = F((X_{t+s})_{s \in \mathcal{B}}, \varepsilon_t).
\label{masterEquation}
\end{equation}

 \cite{doukhan2007fixed} ensures the existence and uniqueness of such solutions giving contraction hypothesis on $F$. Here, we just suppose to have one (strongly) stationary random field $(X_t)_{ t \in \mathbb{Z}^\kappa}$ verifying this equation and we do not assume the uniqueness.
 
Thereafter, we denote by $\mu_X$ the stationary distribution of one marginal random variable $X_{t}$. We assume that laws $\mu_X$ and $\mu_\varepsilon$ are stable by F. It means that, if $(X_{t})_{t \in \mathcal{B}}$ is drawn with marginal distribution $\mu_X$ and $\varepsilon$ is drawn with distribution $\mu_\varepsilon$, then the random variable $F((X_{s})_{s \in \mathcal{B}}, \varepsilon)$ follows the distribution $\mu_X$. 

For $s\in \mathcal{B}$, $X_{t+s}$ depends on $X_{t}$, but $X_{t}$ also depends on $X_{t+s}$. Therefore, it is no longer possible to describe $(X_t)_{ t \in \mathbb{Z}^\kappa}$ as a result of a martingale process. This is why, we call $(X_t)_{ t \in \mathbb{Z}^\kappa}$ a non-causal random field.

\begin{rem}
It is important to note that the evolution of the random field is defined by a local equation. This is similar to settings of auto-regressive times series in the uni-dimensional case. This is why, our framework can be seen as a generalization of stationary Markov chains. Indeed when $\delta = \kappa = 1$ and $\mathcal{B} =\{-1\}$, $(X_t)_{t  \in \mathbb{Z}}$ is an homogeneous and stationary Markov chain.
\end{rem}

\subsubsection{Contraction hypothesis}

Numerous concentration inequalities for Markov chains and other martingales rely on contraction hypotheses (e.g. \cite{dedeckerdeviation2015,dedeckerDeviation2019,alquier2019exponential}).
\\
Let's consider a $\mathcal{X}$-valued Markov chain $(X_t)_{t \in \mathbb{Z}}$ given by
\[
X_t = F(X_{t-1}, \varepsilon_t),
\;
\text{where $\varepsilon_t$ are i.i.d. random variables.}
\]
The classical contraction hypothesis requires that it exists $\rho < 1$ such that for all $y,y' \in \mathcal{X}$,
\begin{equation}
    \esp \left( \| F(y,\varepsilon) - F(y^\prime,\varepsilon) \| \right) \leq \rho \|y-y'\|.
    \label{equ:MC_contraction}
\end{equation}

It is relatively easy to extend this condition to a non-causal random field defined by Equation $\eqref{masterEquation}$. It would be the following condition.
\paragraph{Absolute contraction.}
It exists $(\lambda_t)_{t \in \mathcal{B}}$, such that $\rho := \sum_{t \in \mathcal{B}} \lambda_t < 1$, and for any tuples $\mathcal{Y} = (y_{t})_{t \in \mathcal{B}}$ and $ \mathcal{Y}' = (y_{t}^\prime)_{t \in \mathcal{B}}$ indexed by $\mathcal{B}$.

\begin{align}
     \esp \left( \|F(\mathcal{Y}, \varepsilon) - F(\mathcal{Y}^\prime, \varepsilon)  \| \right) \leq \sum_{t \in \mathcal{B}} \lambda_t \| y_{t} - y_{t}^\prime\|.
\label{stdContraction}
\end{align}

However, this condition, that we called \emph{absolute contraction}, seems to strong for many practical purposes. Indeed, practitioners tend to avoid using absolutely contracting recurrent neural networks, which often lead to a problem called catastrophic forgetting \cite{6707047}. Therefore, it is impossible to use \emph{absolute contraction} when data are generated by such models. Moreover, absolute contraction is difficult to ensures in practice.

We therefore want to relax this condition. To that end, we consider that the contraction is only verified for a $m$ order moment. This leads to the following hypothesis. \\

\paragraph{Weak contraction hypothesis $(\bf{H_1^m})$.} 

\textit{There exist $m \in \mathbb{N}$ and $(\lambda_t)_{t \in \mathcal{B}} \in [0, 1]^{\mathcal{B}}$ such that, for all tuples of random variables $(Y_t)_{t \in \mathbb{Z}^\kappa}, (Y_t^\prime)_{t \in \mathbb{Z}^\kappa}$ with marginal distribution $\mu_{X}$ and i.i.d. random fields $\pmb{\varepsilon} = \inovField$ with product distribution $\mu$.}
\begin{itemize}
    \item $
    \sum\limits_{t \in \mathcal{B}}  \lambda_t < 1
    \text{ (we denote thereafter } \rho = \sum\limits_{t \in \mathcal{B}} \lambda_t),
    $
    \item $
    \forall t \in \mathbb{Z}^\kappa, \|F((Y_{t+s})_{s \in \mathcal{B}}, \varepsilon_t) - F((Y_{t+s}^\prime)_{s \in \mathcal{B}}, \varepsilon_t)\| \text{ and } \|Y_{t} - Y_{t}^\prime\|
    \text{ admit moment at order m,}
   $
   \item$
   \displaystyle
    \forall t \in \mathbb{Z}^\kappa, \|F((Y_{t+s})_{s \in \mathcal{B}}, \varepsilon_t) - F((Y_{t+s}^\prime)_{s \in \mathcal{B}}, \varepsilon_t)  \|_m \leq \sum_{s \in \mathcal{B}} \lambda_s \| Y_{t + s} - Y_{t + s}^\prime \|_m.
   $
\end{itemize}

 We emphasize that $(\bf{H_1^m})$ depends on $m$ because the contraction concerns only the $m-$moment and does not require the function F to be contracting (Equation \eqref{stdContraction}). One benefit of the weak contraction hypothesis is that it should be possible to test it, whereas the absolute contraction cannot be checked on empirical data if the function $F$ is unknown.
 
 However, weak contraction hypothesis leads to weaker results than absolute contraction. Therefore we also use the following compromise: When m goes to $\infty$, $(\bf{H_1^m})$ becomes \\
 
\paragraph{Uniform contraction hypothesis $(\bf{H_1^{\infty}})$.}
\begin{itemize}
    \item \textit{We denote by $\displaystyle \dm{\mathcal{X}} = \underset{y,y^\prime \in \mathcal{X}}{\sup}{\| y - y^\prime \|}$ the diameter of $\mathcal{X}$ and assume that $\dm{\mathcal{X}} < \infty$.}
    \item \textit{We assume it exists $(\lambda_t)_{t \in \mathcal{B}} \in [0, 1]^{\mathcal{B}}$ such that $\sum\limits_{t \in \mathcal{B}}  \lambda_t < 1$.}
    \item \textit{Then, for all $\mathcal{Y} = (y_{t})_{t \in \mathcal{B}} \; \text{and} \; \mathcal{Y}' = (y'_{t})_{t \in \mathcal{B}}$ tuples from $\mathcal{X}^\mathcal{B}$ and for all $e$ in $E$.}
    \begin{equation*}
        \|F(\mathcal{Y}, e) - F(\mathcal{Y}^\prime, e)\| \leq \sum_{s \in \mathcal{B}} \lambda_s \underset{y_s,y_s^\prime \in \mathcal{X}}{\sup} \| y_s - y_s^\prime \| = \rho \dm{\mathcal{X}}.
    \end{equation*}
\end{itemize}

Uniform contraction condition $(\bf{H_1^\infty})$ is still a weaker hypothesis than absolute contraction hypothesis (Equation \eqref{stdContraction}). Indeed, if the diameter of $\mathcal{X}$ is finite, absolute contraction hypothesis (see Equation \eqref{stdContraction}) implies $(\bf{H_1^\infty})$. 

\subsubsection{Coupling hypothesis} 
We introduce a coupling hypothesis similar to one use in \cite{dedeckerdeviation2015}. It controls the moment of the difference between two independents variables following the same distribution $\mu_X$.\\

\paragraph{Weak coupling hypothesis $(\bf{H_2^m})$.}
\textit{Let $m \in \mathbb{N}$. We suppose that for two $\mathcal{X}$-valued independent random variables X, Y with marginal distribution $\mu_X$, it exists a constant $\mathbb{V}_m$ such that}
    \begin{equation*}
        \| Y - X  \|_m \leq \mathbb{V}_m.
    \end{equation*}

This hypothesis is immediately verified as soon as $\dm{\mathcal{X}} \leq \infty$ (because $\forall m, \mathbb{V}_m < \dm{\mathcal{X}}$). Nevertheless, we point out that the quantity $\mathbb{V}_m$ plays a important role in our concentration inequality result and may be significantly smaller than $\dm{\mathcal{X}}$.
    
As in the previous subsection, we provide an alternative hypothesis for $m=\infty$.

\paragraph{Uniform coupling hypothesis $(\bf{H_2^{\infty}})$.}
\textit{We suppose that for two $\mathcal{X}$-valued random variables X, Y with marginal distribution $\mu_X$, it exists a constant $\mathbb{V}_\infty$ such that}
    \begin{equation*}
        \| Y - X  \|_\infty \leq \mathbb{V}_\infty.
    \end{equation*}
   
 In this case, $(\bf{H_2^\infty})$ implies $(\bf{H_2^m})$ for every $m \in \mathbb{N}$. It remains true that $(\bf{H_2^\infty})$ is immediately verified as soon as the process $(X_t)_{t \in \mathbb{Z}^\kappa}$ is bounded. Indeed, $\| Y - X  \|_\infty = \dm{\mathcal{X}}$ and in this case $\mathbb{V}_\infty = \dm{\mathcal{X}}$.

\subsubsection{Examples}
We provide below some examples of non-causal random fields.\\

\textit{Non-causal linear fields.}

Under our framework, the simplest possible non-causal random fields is the bidirectional linear model. In this case $\kappa= 1$, and we have $\alpha_{-1}, \alpha_1$ such that

\[F(X_{t-1}, X_{t+1}, \varepsilon_t) = \alpha_{-1} X_{t-1} + \alpha_{1} X_{t+1} + \varepsilon_t.\]

Where the $\varepsilon_t$ are a Gaussian white noise of variance $\sigma^2$. 
In this setting, the condition $(\bf{H_1^m})$ is satisfied if $\alpha_{-1} + \alpha_{1}  < 1$.

\begin{rem}
This is equivalent, when $\alpha_1 \not = 0$ to a classical AR(2) model.
\end{rem}

Bidirectional linear field can be generalize to linear random fields

\begin{equation*}
    X_t = F((X_{t+i})_{i \in \mathcal{B}}, \varepsilon_t) = \sum_{s \in \mathcal{B}} \alpha_s X_{t+s} + \varepsilon_t.
\end{equation*}
In any case, the assumption $(\bf{H_1^m})$ is satisfied providing $\sum_{s \in \mathcal{B}} \alpha_s < 1$.

\textit{Finite LARCH random fields.}Finite LARCH(n) random fields \cite{surgailis_quadratic_2008} are defined by 

\begin{equation*}
    X_t = F((X_{t-i})_{i \in [1,n]}, \varepsilon_t) = \varepsilon_t \left( \alpha_0 + \sum_{i =1}^n \alpha_i X_{t-i} \right).
\end{equation*}
 They can be generalized to non-causal LARCH(n) fields defined by
 \begin{equation*}
    X_t = F((X_{t+s})_{s \in \mathcal{B}}, \varepsilon_t) = \varepsilon_t \left( \alpha_0 + \sum_{s \in \mathcal{B}} \alpha_s X_{t+s} \right).
\end{equation*}
 
 In this case, a sufficient condition to fulfill $(\bf{H_1^m})$ is $\displaystyle \| \varepsilon_t \|_{\infty} \sum_{s \in \mathcal{B}} \alpha_s < 1$.\\

\textit{Finite ARCH random fields.}
ARCH models are widely used in econometric and can be easily extended to the non-causal case. Here, we propose a Bi-ARCH(1,1) model defined by 

\[
X_t = F(X_{t-1},X_{t+1}, \varepsilon_t) = \varepsilon_t \sqrt{ \alpha_{-1} X_{t-1} + \alpha_{1} X_{t+1} + \beta},
\;
\text{where $\varepsilon$ is a real random variable.}
\]
In this case, $(\bf{H_1^m})$ is satisfied providing $\|\varepsilon\|_{\infty} (\alpha_1 +\alpha_{-1}) < 1$. $(\bf{H_2^m})$ is satisfied if $\varepsilon$ is bounded and the process stationary.
\\
Those process can be extend to the multidimensional case, such process verify 
\begin{equation*}
    X_t = F((X_{t+s})_{s \in \mathcal{B}}, \varepsilon_t) = \varepsilon_t \sqrt{ \alpha_0 + \sum_{s \in \mathcal{B}} \alpha_s X_{t+s}^2 }.
\end{equation*}\\

\textit{Bidirectional RNN.}

Bidirectional recurrent neural networks (BRNN) have been used  in Natural Language Processing. They have many applications from text translation to part of speech tagging and speech recognition. 
Here, we present the formal version of a single-layer bidirectional neural network with a white noise $\varepsilon_t$.

\[
F(\mathcal{X}_t ,\varepsilon_t)  =  f\Big( A \mathcal{X}_t \Big)+ \varepsilon_t.
\]

Where $A$ is a $p \times 2k$ matrix, $f$ is an activation function and $\mathcal{X}_t$ is the 2k vector  $(X_{t-k} \dots, X_{t-1}, X_{t+1},\dots, X_{t+k})$.
\\
We suppose that the activation function is 1-Lipschitz. This is the case for most activation function (sigmoid, tanh, RELU, softmax). There is also an operator norm $\|\|_{op,m}$ associated with the norm $\|\|_m$. \\
With this condition, the contraction condition $(\bf{H_1^m})$ is verified as soon as
\[\|A\|_{op,m}  < 1.\]

If the white noise is bounded, the condition $(\bf{H_2^{\infty}})$ is verified. Instead, if it is subgaussian, we only have $(\bf{H_2^m})$.

\subsection{Function of interest $\Phi$ and the statistic $\SI$}
\label{statSI}
Throughout this article, we focus on a function $\Phi : \mathcal{X}^{\Bar{\mathcal{B}}} \mapsto \mathbb{R}$ defined on a small neighborhood
\begin{equation}
    \Bar{\mathcal{B}} = \mathcal{V}(\Bar{\delta},0) = \prod_{i=1}^{\kappa} [-\Bar{\delta}_{i}, \Bar{\delta}_i].
\end{equation}
And we denote by $n_{\Bar{B}}$ the cardinal of $\Bar{\mathcal{B}}$. It is important to clarify that definitions of $\mathcal{B}$ and $\Bar{\mathcal{B}}$ are  slightly different, the point $\{0\}$ is included in the set $\Bar{\mathcal{B}}$ and not in $\mathcal{B}$. Whereas definitions of $n_B$ and $n_{\Bar{B}}$ are comparable, indeed $n_B = card(\mathcal{B}) + 1 $ and $n_{\Bar{B}} = card(\Bar{\mathcal{B}})$.
\\
Then, for a given subset $\mathcal{I}$ of indices, we define the statistic $\SI$
\begin{equation}
    \SI= \sum_{s  \in \mathcal{I}} \Phi((X_{s+t})_{t \in \Bar{\mathcal{B}}})\,.
    \label{eq:SI_model}
\end{equation}

We first recall that the process is \emph{strongly} stationary, thus $\esp[\Phi((X_{s+t})_{t \in \Bar{\mathcal{B}}})] = \esp[\SI] $ is well defined. Our goal is to control the difference between $\SI$ and $\esp[\SI]$, which is the deviation of $\SI$.We introduce below some hypotheses concerning this function $\Phi$.\\

\textbf{Lipschitz separability hypothesis $(\bf{H_3})$.}
\textit{We assume that $\Phi$ is L-Lipschitz separable, i.e., there exists a constant $L > 0$ such that}
 \begin{equation}
 \textit{
 For all $(u_t)_{t \in \Bar{\mathcal{B}}}$, $(v_t)_{t \in \Bar{\mathcal{B}}}$ tuples of $\mathcal{X}^{\Bar{\mathcal{B}}}$}, \; \|\Phi((u_t)_{t \in \Bar{\mathcal{B}}}) - \Phi((v_t)_{t \in \Bar{\mathcal{B}}})\| \leq L \sum_{t \in \Bar{\mathcal{B}}} \|u_t - v_t\|.
\end{equation}

This hypothesis is close to the condition proposed by \cite{dedeckerdeviation2015} in the causal dependent case. Moreover, recent works (see \cite{virmauxlipschitznodate}) suggest that such hypothesis is suitable to deal with deep learning models and provide algorithm to estimate the Lipschitz constant.
\\
To simplify, we assume throughout the article that $L=1$ but setting $L \ne 1$ would only add a multiplicative factor in our results.\\

\textbf{Bound for $\Phi$ $(\bf{H_4})$.}
\textit{We assume that $\Phi$ is bounded.}
\begin{equation}
    \forall x \in \mathcal{X}^{\Bar{\mathcal{B}}}, |\Phi(x)| \leq M.
    \label{phiBound}
\end{equation}


\section{Main Results}
\label{Sec:Main}
\subsection{Concentration inequality within i.i.d. assumption}
\label{learningTh}

 Numerous concentration inequalities have been established to control the deviation of a random variables. Hoeffding's and McDiarmid's inequalities, proposed respectively in \cite{hoeffding} and \cite{mcdiarmid1989}, are among the most widely used in machine learning. Their classical versions are based on the i.i.d. assumption. 
\\ 
Below, we recall the Hoeffding's inequality.

\begin{theorem}[Hoeffding's inequality] 
Let $X_1, \dots, X_n$ be independent real random variables and $\SI = \sum_{i=1}^n X_i$. \\
We suppose that there exist two tuples $(a_1, \dots, a_n) \; \text{and} \; (b_1, \dots, b_n)$ such that with probability 1,
\begin{equation}
    \forall i \in [1,n], a_i \leq X_i \leq b_i.
\label{equ:strong_difference_bound}
\end{equation}
Then
\begin{equation}
    \pr \left( \SI - \esp \left[ \SI \right] \geq \varepsilon \right) \leq \exp \left( \frac{-2\varepsilon^2}{\sum_{i=1}^n (b_i-a_i)^2} \right).
\end{equation}
\label{ineq:Hoeffding}
\end{theorem}

In the field of statistical learning, Hoeffding's and McDiarmid's have lots of applications (\cite{massart2007concentration, boucheron2013concentration} give examples). 
\\
In this article, we aim to prove a Hoeffding-type inequality in the non-causal setting defined in Section \ref{Sec:Model}.


\begin{rem}
Other types of concentration inequalities (Bernstein, von Bahr-Esseen) could have been proved in the same context using the same approach. Indeed, it should also be possible, as in the i.i.d. case, to derive extra inequalities from the exponential inequality stated in Lemma \ref{mDev}. 

Nevertheless, we focus on Hoeffding's inequality for its simplicity and its use in many applications.
\end{rem}

\subsection{Concentration inequalities and expected deviation bounds for non-causal random fields}

We state below simplified versions of our results. Full theorems may be found in Section \ref{fullResults}.

\begin{theorem}[Concentration inequality, uniform contraction case]
    Let $n_B = \card{\mathcal{B}} + 1$ and $n_{\Bar{B}} =  \card{\Bar{\mathcal{B}}}$. If $(\bf{H_1^\infty})$, $(\bf{H_2^\infty})$ and $(\bf{H_3})$ are verified, there is a constant $A$ such that, for $\varepsilon >  2  n_{\Bar{B}} \mathbb{V}_{\infty} $,
    \begin{align*}
     \pr \left( |\SI - \esp \left[ \SI \right]  | \geq \varepsilon \right) \leq 2 \exp \left(\frac{ -2\left(  \varepsilon - 2  n_{\Bar{B}} \mathbb{V}_{\infty}  \right)^2  }{\left( n_{\Bar{B}} \mathbb{V}_{\infty} \right)^2 \left( 1 + A n_{\Bar{B}} n_B^3 \kappa!^2 \left\lceil \ln(n) \right\rceil^\kappa n \right)} \right).
    \end{align*}
\label{simplConcentrationStrong}
With $A$ such that $\Upsilon(d)^2 d^\kappa \leq A (\kappa!)^2 \left\lceil \ln(n) \right\rceil^\kappa$, where $\Upsilon$ is the function defined in Lemma \ref{lem:technical} and can be bounded independently of $n_{\Bar{B}}$, $m$ or $n$.

\end{theorem}

Constant $A$ does not depend on $n_{B}$, $n_{\Bar{B}}$ or $n$ and is explicit it may be found in Theorem \ref{infConcentration}.
\\
The dominant term in the denominator is $A \mathbb{V}_{\infty}^2 n_{\Bar{B}}^3 n_{\mathcal{B}}^3 \kappa!^2 \left\lceil \ln{n} \right\rceil^\kappa n$, its asymptotic behavior is slightly less good than in the i.i.d. Hoeffding's inequality. Indeed, it increases as $\mathcal{O}(n(\ln{n})^\kappa)$ instead of $\mathcal{O}(n)$.
\\
Moreover, this term is strongly impacted by the dimension of the random field $\kappa$. However, this limitation is not a problem in practice, because, in most practical cases, $\kappa \in{1,2}$.
\\
Other important factors are parametric dimensions of our model, represented by $n_{B}$ and $n_{\Bar{B}}$. Quite logically, the quality of the inequality decreases when the number of variables to control increases. Our inequality is quite sensible to those factors and consequently is not a suitable result for high dimension estimation ,i.e., when $n$ is smaller than $n_{B}$ and $n_{\Bar{B}}$. We recall that for a simple Markov chain both of these terms would be equal to 2.

Finally, the condition $\varepsilon \geq  2  n_{\Bar{B}} \mathbb{V}_{\infty}$ does not seem restrictive for applications as we will show in subsection \ref{subsec:Learning}.

We now stand a simplified version of our concentration inequality for the weaker assumptions $(\bf{H_1^m})$ and $(\bf{H_2^m})$.

\begin{theorem}[Concentration inequality, weak contraction case]
\label{simplConcentrationWeak}
     Let $n_B = \card{\mathcal{B}} + 1$ and $n_{\Bar{B}} =  \card{\Bar{\mathcal{B}}}$. If $(\bf{H_1^m})$, $(\bf{H_2^m})$, $(\bf{H_3})$ and $(\bf{H_4})$ are verified, there are constants $A$,$B$,$C$,$D$,$E$,$F$,$H$ such that, for $\varepsilon \geq  2 F \left(n_{\Bar{B}} n_B\right)^3 \kappa! \left\lceil \ln(n) \right\rceil^{2 \kappa} n^{\frac{2}{m}}$.
\begin{align*}
    \pr \left( |\SI - \esp \left[ \SI \right]  | \geq \varepsilon \right) \nonumber 
    & \leq 2 \exp \left(  \frac{-  2 \left( \frac{\varepsilon}{2} - F \left(n_{\Bar{B}} n_B\right)^3 \kappa! \left\lceil\ln(n) \right\rceil^{2 \kappa} n^{\frac{2}{m}} \right)^2}{ \left(H n_{\Bar{B}}  n^{\frac{2}{m}} \right)^2 \left(1 + E n_{\Bar{B}} n_B^3 (\kappa!)^2 \left\lceil \ln(n) \right\rceil^\kappa n \right) }\right) \\
    & + \frac{\rho^m}{n} \left( 2 n_B n_{\Bar{B}} C \left\lceil \ln(n) \right\rceil^\kappa + \left( \frac{D}{n_B^3 n_{\Bar{B}}^2 \Upsilon(d) \ln(n)^{2\kappa}} \right)^m \right).
\end{align*}
These constants are 
\begin{itemize}
    \item $A=\frac{1 + \frac{\ln(\rho^{-1})}{\kappa! n_B} + \ln(\rho^{-1}) \frac{ \left(\frac{\kappa - 1}{e} \right)^{\kappa - 1}}{(\kappa - 1) !}}{\ln(\rho^{-1})^\kappa}$, then $\Upsilon(d) \leq A \kappa!$, where $\Upsilon$ is the function defined in Lemma \ref{lem:technical} and can be bounded independently of $n_{\Bar{B}}$, $m$ or $n$.
    \item $B= \frac{1}{\rho^2} + \frac{1}{\rho \left(n_{\Bar{B}} n_B  \right)^{\frac{1}{m}}} + \frac{2M}{ \Upsilon(d) \mathbb{V}_m \left(n_{\Bar{B}} n_B \right)^2} $.
    \item $C= \left\lceil \frac{1 - \frac{1}{m}}{\ln(\rho^{-1})}  \right\rceil^\kappa $, then $d^\kappa \leq C \left\lceil \ln(n) \right\rceil^\kappa$.
    \item $D = \frac{\rho \ln(\rho^{-1})^{2\kappa}}{2 \left(1 - \frac{1}{m} \right)^{2 \kappa}}$.
    \item E such that $\Upsilon(d)^2 d^\kappa \leq E (\kappa!)^2 \left\lceil \ln(n) \right\rceil^\kappa$.
    \item $F=2BC^2A \mathbb{V}_m$.
    \item $H = \frac{\mathbb{V}_m}{\rho}$
\end{itemize}

\end{theorem}

Similar comments on the dimension $\kappa$ and parameters $n_{B}$, $n_{\Bar{B}}$ to those concerning previous theorem can be done.
\\
$(\bf{H_1^m})$ and $(\bf{H_2^m})$ are fairly weaker assumptions than $(\bf{H_1^\infty})$ and $(\bf{H_2^\infty})$ and thus lead to degraded concentration inequalities. Indeed, in the exponential term, the denominator is asymptotically dominated by $\mathcal{O}\left( n^{1+\frac{4}{m}} (\ln(n))^\kappa \right)$ instead of $\mathcal{O}\left( n (\ln(n))^\kappa \right)$ under assumptions $(\bf{H_1^\infty})$ and $(\bf{H_2^\infty})$ and $\mathcal{O}\left( n  \right)$ under i.i.d. assumption.
\\
We point out that this theorem only offers interesting results when $m > 4$ (the reason why appears more clearly in Theorem \ref{ErrGenTail}).
\\
The two extra additive terms decrease faster than the main term and are not dominant for applications as we will show in subsection \ref{subsec:Learning}.

These two theorems lead to the next corollaries for the expected deviation.

\begin{cor}[Simplified expected deviation bound, uniform contraction case]
   We assume $(\bf{H_1^\infty})$, $(\bf{H_2^\infty})$ and $(\bf{H_3})$, with the same constant $A$ occurring in Theorem \ref{simplConcentrationStrong}. It holds
\begin{equation*}
    \esp[|\SI - \esp \left[ \SI \right]|] \leq n_{\Bar{B}} \mathbb{V}_{\infty} \left( 2 + \sqrt{\frac{\pi}{2} \left( 1 + A n_{\Bar{B}} n_{B}^3 (\kappa!)^2 \left\lceil \ln(n) \right\rceil^\kappa n \right)} \right).
\end{equation*}
\label{simplStrongMomentIneq}
\end{cor}

\begin{cor}[Simplified expected deviation bound, weak contraction case]
    We assume $(\bf{H_1^m})$, $(\bf{H_2^m})$, $(\bf{H_3})$ and $(\bf{H_4})$, with the same constants $A$,$B$,$C$,$D$,$E$,$F$,$H$ occurring in Theorem \ref{simplConcentrationWeak}. It holds
\begin{align*}
    & \esp[|\SI - \esp \left[ \SI \right]|] \nonumber \\
    & \leq  2 n_{\Bar{B}} H n^{\frac{2}{m}} \sqrt{\frac{\pi}{2} \left( 1 + E n_{\Bar{B}} n_{B}^3 (\kappa!)^2 \left\lceil \ln(n) \right\rceil^\kappa n \right)} + 2 F \left(n_{\Bar{B}} n_B\right)^3 \kappa!  \left\lceil \ln(n) \right\rceil^{2 \kappa} n^{\frac{2}{m}} \nonumber \\
    & + 2 \rho^m M \left( 2 n_B n_{\Bar{B}} C \left\lceil \ln(n) \right\rceil^\kappa + \left( \frac{D}{n_B^3 n_{\Bar{B}}^2 \Upsilon(d) \ln(n)^{2\kappa}} \right)^m \right).
\end{align*}
\label{simplWeakMomentIneq}
\end{cor}

\begin{rem}
 Hypotheses $(\bf{H_1^m})$ and $(\bf{H_2^m})$ do not require random variables $(X_t)_{t \in \mathbb{Z}^\kappa}$ to be bounded, unlike $(\bf{H_1^\infty})$ and $(\bf{H_2^\infty})$. Even tough $\dm{\mathcal{X}} < \infty$, it may be advantageous to use $(\bf{H_1^m})$ and $(\bf{H_2^m})$ instead of $(\bf{H_1^\infty})$ and $(\bf{H_2^\infty})$ when the bound $\dm{\mathcal{X}}$ is too high.
\end{rem}

\subsection{Comparison with other concentrations inequalities}

In this part, we compare our results with others concentration inequalities and their applications in learning theory. Many concentration inequalities have been established for wide variety of settings. We focus on the five following settings.

\paragraph*{Independent and identically distributed} The simplest possible case is the case where random variables are pairwise independent. We have already presented the classical Hoeffding's inequality. McDiarmid's one is similar but is slightly more general, as it does not concern a sum of the random variables $\sum_{i=1}^n X_i$ , but a general random function of random variables $\phi(X_1,\dots, X_n)$. To a certain extent, our concentration inequalities are a mix between Hoeffding's and McDiarmid's inequality, as we propose a result for a sum of such functions.

Requirements for Hoeffding's inequality (see Equation $\eqref{equ:strong_difference_bound}$) or its equivalent for McDiarmid's inequality, are \emph{uniform} in the sense of \cite{kutin2002extensions} (uniform bounded difference assumption). However, some attempts \cite{kutin2002extensions,combesExtension2015,wudistributiondependent2018} have been made to relax this condition. Instead of assuming such conditions with probability 1, they may be only required within a probability $1-\rho$ (with $\rho$ small). However, relaxing those conditions decreases the quality of the bound.
This can be compared with the difference between conditions $(\bf{H_1^{\infty}})$ and $(\bf{H_1^m})$. Indeed, the moment condition implies that the application F is a contracting mapping within a high probability. 

\paragraph*{Markov chain.} A first way to model dependence in a causal case is to describe the process as a Markov chain, where the new values $X_{t}$ verify an equation of type $X_t = F(X_{t-1}, \varepsilon_t)$ for an (i.i.d) innovation $\varepsilon_t$. We drew a lot from this approach, our non-causal main Equation \eqref{masterEquation} has the same form as a classical Markov chain causal equation.

In order to establish concentration inequalities on Markov chain, further assumptions on the behavior of the function F are needed. For example, classical hypotheses include mixing time and spectral gap conditions \cite{paulin2015concentration}. However, these conditions are hard to translate in a non-causal and multi-dimensional case. Therefore, we use a similar approach than \cite{dedeckerdeviation2015, alquier2019exponential} and introduce a contraction hypothesis.
Nevertheless, as far as we know, our contraction hypotheses $(\bf{H_1^{\infty}})$ and $(\bf{H_1^{m}})$ are weaker than the absolute contraction hypothesis used in previous works (e.g. \cite{dedeckerdeviation2015}).  
That is one reason why the rate of convergence obtained by \cite{dedeckerdeviation2015} is better than our own. They get a convergence rate in $\mathcal{O}(\sqrt{n})$ instead of our the $\mathcal{O}(\sqrt{\ln(n)n})$ of our Corollary \ref{simplStrongMomentIneq}.

\paragraph*{Weak dependence for causal times series.}
Another possibility is to use weak dependence (see \cite{dedecker2007weak}) to model the dependence in time series. They can be applied to more general processes, because they take into account long-range dependence. However, weak dependence conditions are non-local, and therefore harder to verify. In causal case, they lead to similar speed of convergence than our results under uniform contraction hypothesis. For instance, they are used by \cite{alquier2012model} for model selection to obtain a convergence rate asymptotically dominated by $\mathcal{O}(\sqrt{\ln^{\frac{5}{2}}(n)n})$.


\paragraph*{Weak dependence for non-causal times series.} 
For non-causal times series, many results have been established using the notion of weak dependence \cite{dedecker2007weak}. This notion replace our contraction condition $(\bf{H_1^{\infty}})$ or $(\bf{H_1^{m}})$. However, it is hard to compare our setting with those results because their concentration inequalities are not Hoeffding or Mc-Diarmid type inequalities. In particular, they introduce an auto-covariance term which does not appear in our concentration inequalities.

\paragraph*{Non-causal random fields.}

Few results are similar to our for general random field. The usual setting used to deal with non-causal random fields is the Markovian random fields setting, where random variables are dispersed on a graph, instead of lattice. There are, at our knowledge, no similar concentration results holding for general markovian random fields.\\
One interesting sub-case of markovian random fields is the Ising model, often used in physics and imagery \cite{mccoy2014two}. Ising models can be viewed as a special case of our model, where $\mathcal{X} = \{-1,1\}$ and $\overline{\mathcal{B}}=\{-1,1\}^{\kappa}$. The transition is governed by the following equation

\begin{equation*}
F((X_{t+s})_{s \in \mathcal{B}}, \varepsilon_t) = 2  \mathbbm{1}\left(\varepsilon > \frac{\exp( \beta \sum_{s \in B} X_{t+s}  )}{\exp( \beta \sum_{s \in B} X_{t+s}) + \exp( - \beta \sum_{s \in B} X_{t+s}  )}\right)  - 1, \end{equation*}

where $\varepsilon_t$ follow an uniform distribution in $(0,1)$. Some concentration results have been established involving coupling conditions \cite{chazottes2007concentration}, it leads to polynomial (instead of exponential) concentration inequalities. It is however hard to compare with our results, as, even if $(\bf{H_1^{\infty}})$ is not fulfilled for Ising model, $(\bf{H_1^m})$ seems reasonable but hard to verify.

\subsection{Application to Learning}
\label{subsec:Learning}
In this subsection, we provide an application of our results to learning theory. We adapt the approach of \cite{bartlett_model_2000}, \cite{lindvall2002lectures} and \cite{Lugosi2002} with our framework, to get an oracle bound for the model selection problem. 

\subsubsection{Application to the completion problem}

A classical problem in learning theory is the completion problem (both in regression or classification). The objective is to predict $X_t$ using its neighbours on a $\kappa$ dimensional lattice $(X_t)_{t \in \Bar{\mathcal{B}}}$. The prediction is given by a model $\model$
\begin{align*}
    \widehat{x}_s = \model((x_{s+t})_{t \in \Bar{\mathcal{B}} \backslash \{0\} }).
\end{align*}

This completion problem may seem trivial, but it is actually the backbone of many NLP challenges. Indeed, predicting words is often used as a primary task to train the encoder in an encoder-decoder scheme. It creates a language model, i.e., a distribution of probability of a word in a text. This language model is then used for more complex task (translation, text segmentation, question answering etc.).

We also introduce a cost function $c: \mathcal{X}^2 \mapsto \mathbb{R}$. $c(\widehat{x}_s,x_s)$ quantifies the error made when the considered algorithm predict $\widehat{x}_s$ instead of $x_s$. In this case, the function of interest $\Phi$ introduced earlier corresponds to the cost of $\model$ on a sample
\begin{equation*}
    \Phi((X_{s+t})_{t \in \Bar{\mathcal{B}}}) = c(\model((X_{s+t})_{t \in \Bar{\mathcal{B}} \backslash \{0\} }), X_s).
\end{equation*}
And 
\begin{equation*}
    \SI = \sum_{s  \in \mathcal{I}} \Phi((X_{s+t})_{t \in \Bar{\mathcal{B}}}) = \sum_{s  \in \mathcal{I}} c(\model((X_{s+t})_{t \in \Bar{\mathcal{B}} \backslash \{0\} }), X_s).
\end{equation*}

\begin{rem}
We emphasize that $\mathcal{B}$ and $\Bar{\mathcal{B}}$ may be different in practice. Indeed, in practice, we can not know the size of $\mathcal{B}$. We may also want to try a simpler model (with  $n_{\Bar{\mathcal{B}}} < n_{\mathcal{B}}$) to reduce overfitting.
\end{rem}

\begin{rem}
The cost function $c: \mathcal{X}^2 \mapsto \mathbb{R}$ can be chosen by the user. Classical choices include quadratic distance, cross entropy, Kullback-Leibler divergence, etc. We point out that it is always possible to bound this cost function by a constant and then to divide this function by this constant.
\\
Therefore, we could assume that $M = 1$ in hypothesis $(\bf{H_4})$ without loss of generality.  
\end{rem}

\begin{figure}[!h]
    \definecolor{ffxfqq}{rgb}{1,0.4980392156862745,0}
    \definecolor{zzttqq}{rgb}{0.6,0.2,0}
    \definecolor{xdxdff}{rgb}{0.49019607843137253,0.49019607843137253,1}
    \definecolor{yqyqyq}{rgb}{0.5019607843137255,0.5019607843137255,0.5019607843137255}
    \definecolor{qqzzff}{rgb}{0,0.6,1}
    \centering
    \begin{tikzpicture}[line cap=round,line join=round,>=triangle 45,x=0.75cm,y=0.75cm]
    \clip(-3,-3) rectangle (14,10);
    \fill[line width=2pt,color=qqzzff,fill=qqzzff,fill opacity=0.1] (0,0) -- (0,5) -- (10,5) -- (10,0) -- cycle;
    \fill[line width=2pt,dash pattern=on 1pt off 1pt,color=yqyqyq,fill=yqyqyq,fill opacity=0.1] (-2,7) -- (12,7) -- (12,-2) -- (-2,-2) -- cycle;
    \fill[line width=2pt,color=zzttqq,fill=zzttqq,fill opacity=0.10000000149011612] (3,7) -- (7,7) -- (7,3) -- (3,3) -- cycle;
    \fill[line width=2pt,dash pattern=on 1pt off 1pt,color=ffxfqq,fill=ffxfqq,fill opacity=0.1] (2,8) -- (8,8) -- (8,2) -- (2,2) -- cycle;
    \draw [line width=2pt,color=qqzzff] (0,0)-- (0,5);
    \draw [line width=2pt,color=qqzzff] (0,5)-- (10,5);
    \draw [line width=2pt,color=qqzzff] (10,5)-- (10,0);
    \draw [line width=2pt,color=qqzzff] (10,0)-- (0,0);
    \draw [line width=2pt,dash pattern=on 1pt off 1pt,color=yqyqyq] (-2,7)-- (12,7);
    \draw [line width=2pt,dash pattern=on 1pt off 1pt,color=yqyqyq] (12,7)-- (12,-2);
    \draw [line width=2pt,dash pattern=on 1pt off 1pt,color=yqyqyq] (12,-2)-- (-2,-2);
    \draw [line width=2pt,dash pattern=on 1pt off 1pt,color=yqyqyq] (-2,-2)-- (-2,7);
    \draw [line width=2pt,color=zzttqq] (3,7)-- (7,7);
    \draw [line width=2pt,color=zzttqq] (7,7)-- (7,3);
    \draw [line width=2pt,color=zzttqq] (7,3)-- (3,3);
    \draw [line width=2pt,color=zzttqq] (3,3)-- (3,7);
    \draw [line width=2pt,dash pattern=on 1pt off 1pt,color=ffxfqq] (2,8)-- (8,8);
    \draw [line width=2pt,dash pattern=on 1pt off 1pt,color=ffxfqq] (8,8)-- (8,2);
    \draw [line width=2pt,dash pattern=on 1pt off 1pt,color=ffxfqq] (8,2)-- (2,2);
    \draw [line width=2pt,dash pattern=on 1pt off 1pt,color=ffxfqq] (2,2)-- (2,8);
    \draw (3.96,8.06) node[anchor=north west] {$\mathcal{V}(\delta,t)$};
    \draw (3.86,7.02) node[anchor=north west] {$\mathcal{V}(\Bar{\delta}),t)$};
    \draw (2.66,-1.1) node[anchor=north west] {Known $X$ Values};
    \draw (2.72,1.18) node[anchor=north west] {Training set $\mathcal{I}$};
    \begin{scriptsize}
    \draw [fill=xdxdff] (5,5) circle (2.5pt);
    \draw[color=xdxdff] (5.22,5.49) node {$X_t$};
    \end{scriptsize}
    \end{tikzpicture}
    \caption{Learning diagram}
    \label{fig:Principe apprentissage}
\end{figure}
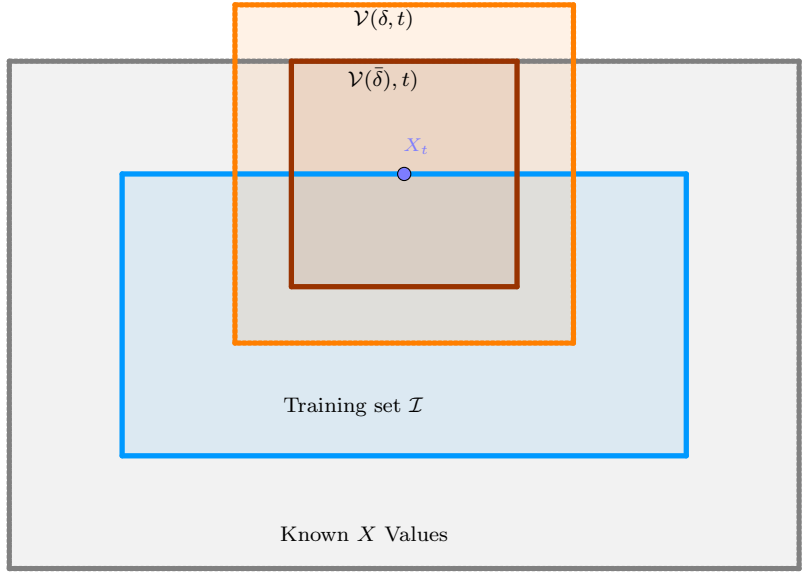

In the Figure \ref{fig:Principe apprentissage}, the red neighborhood $\mathcal{V}(\Bar{\delta},t)$ is used to compute $\widehat{x}_t = \model((x_{t+s})_{s \in \Bar{\mathcal{B}}})$ which approximate $x_t$.
\\
It implies that the set $\mathcal{V}(\Bar{\delta},t)$ is included in the set of known values $X$ (the grey set), for each $t$ in $\mathcal{I}$ (the blue set). 

In the completion problem, we want to control the gap between the theoretical risk of an estimator $\model$, denoted by $\rth{\model}$,

\begin{equation}
    \rth{\model} = \esp[c(\model((X_{s+t})_{t \in \Bar{\mathcal{B}} \backslash \{0\} }), X_s)], 
\end{equation}

and its empirical risk,

\begin{equation}
    \remp{\model} = \frac{1}{n} \sum\limits_{s \in \mathcal{\mathcal{I}}} c(\model((X_{s+t})_{t \in \Bar{\mathcal{B}}\backslash \{0\}}), X_s) \label{empRisk}.
\end{equation}

Immediate application of our results gives tail and moment inequalities for $|\rth{\model} - \remp{\model}|$ under hypotheses $(\bf{H_1^\infty})$, $(\bf{H_2^\infty})$ and $(\bf{H_1^m})$, $(\bf{H_2^m})$.

\begin{theorem}[Tail inequalities]
 Let $n_B = \card{\mathcal{B}} + 1$ and $n_{\Bar{B}} =  \card{\Bar{\mathcal{B}}}$.
 \begin{itemize}
     \item If $(\bf{H_1^\infty})$, $(\bf{H_2^\infty})$ and $(\bf{H_3})$ are verified, there are constants such that, for $\varepsilon >  \frac{2  n_{\Bar{B}} \mathbb{V}_{\infty} }{n} $,
    \begin{align*}
    \pr \left(|\rth{\model} - \remp{\model}| \geq \varepsilon \right) \leq 2 \exp \left(\frac{ -2 n^2 \left(  \varepsilon - \frac{2  n_{\Bar{B}} \mathbb{V}_{\infty}}{n}  \right)^2  }{\left( n_{\Bar{B}} \mathbb{V}_{\infty} \right)^2 \left( 1 + A n_{\Bar{B}} n_B^3 \kappa!^2 \left\lceil \ln(n) \right\rceil^\kappa n \right)} \right).
    \end{align*}
    \item If $(\bf{H_1^m})$, $(\bf{H_2^m})$, $(\bf{H_3})$ and $(\bf{H_4})$ are verified, there are constants such that, for $\varepsilon \geq \frac{2 L_1(n)}{n^{1-\frac{2}{m}}}$,
    \begin{align*}
    &\pr \left(|\rth{\model} - \remp{\model}| \geq \varepsilon \right) 
    \leq 2 \exp \left(  \frac{-  2 n^{2-\frac{4}{m}} \left( \frac{\varepsilon}{2} - \frac{L_1(n)}{n^{1-\frac{2}{m}}}  \right)^2}{ \left(H n_{\Bar{B}} \right)^2 \left(1 + E n_{\Bar{B}} n_B^3 (\kappa!)^2 \left\lceil \ln(n) \right\rceil^\kappa n \right) }\right) + \frac{\rho^m}{n} L_2(n),
\end{align*}
with $L_1(n) = F \left(n_{\Bar{B}} n_B\right)^3 \kappa! \left\lceil\ln(n) \right\rceil^{2 \kappa}$ and $L_2(n) = 2 n_B n_{\Bar{B}} C \left\lceil \ln(n) \right\rceil^\kappa + \left( \frac{D}{n_B^3 n_{\Bar{B}}^2 \Upsilon(d) \ln(n)^{2\kappa}} \right)^m$.
 \end{itemize}
\label{ErrGenTail}
\end{theorem}

\begin{proof}
It is a direct application of Theorems \ref{simplConcentrationStrong} and \ref{simplConcentrationWeak} with $\SI = \sum\limits_{s \in \mathcal{\mathcal{I}}} c(\model((X_{s+t})_{t \in \Bar{\mathcal{B}} \backslash \{0\}}), X_s)$. Moreover, constants involved are the same as in Theorem \ref{simplConcentrationStrong} and \ref{simplConcentrationWeak}.
\end{proof}

If we assume that $\esp[\exp(| \rth{\model} - \remp{\model} |)]$ exists, we get the following useful result.

\begin{cor}
~\begin{itemize}
     \item If $(\bf{H_1^\infty})$, $(\bf{H_2^\infty})$ and $(\bf{H_3})$ are verified. For each $s > 0$,
     \begin{align}
        \esp[\exp(s | \rth{\model} - \remp{\model} |)]
        \leq & \exp \left( \frac{2 s n_{\Bar{B}} \mathbb{V}_{\infty} }{n} + \frac{\left( n_{\Bar{B}} \mathbb{V}_{\infty} s \right)^2 \left( 1 + A n_{\Bar{B}} n_B^3 \kappa!^2 \left\lceil \ln(n) \right\rceil^\kappa n \right)}{8 n^2} \right) \nonumber \\
        & \times \left( 1 + \frac{n_{\Bar{B}} \mathbb{V}_{\infty} s}{n} \sqrt{2 \pi \left( 1 + A n_{\Bar{B}} n_B^3 \kappa!^2 \left\lceil \ln(n) \right\rceil^\kappa n \right)}  \right). 
    \label{strongexpMoment}
    \end{align}
     \item If $(\bf{H_1^m})$, $(\bf{H_2^m})$, $(\bf{H_3})$ and $(\bf{H_4})$ are verified. For each $s >0$,
    \begin{align}
        & \esp[\exp(s | \rth{\model} - \remp{\model} |)] \nonumber \\
        & \leq
        \exp \left( \frac{1}{2 n^{1-\frac{4}{m}}} \left( 4sL(n) + \left(H n_{\Bar{B}} s \right)^2 \left(1 + E n_{\Bar{B}} n_B^3 (\kappa!)^2 \left\lceil \ln(n) \right\rceil^\kappa n \right) \right) \right) \nonumber \\
        &\times \left( 1 + \frac{\exp \left( Ms \right) \rho^m L_2(n)}{n}  
        + \frac{2H n_{\Bar{B}} s}{n^{1-\frac{2}{m}}} \sqrt{2 \pi \left(1 + E n_{\Bar{B}} n_B^3 (\kappa!)^2 \left\lceil \ln(n) \right\rceil^\kappa n \right)}
        \right).
        \label{weakexpMoment}
    \end{align}
\end{itemize}
\label{expMoment}
\end{cor}

The proof can be found in the appendix \ref{proofExpMoment}.

\subsubsection{Model selection}

In this subsection, we present a model selection result for a finite set of models. It is an extension of well known results to our framework (\cite{bartlett_model_2000}). It should also be possible to get results for a infinite set of learning function using Vapnik-Chervonenkis dimension \cite{blumer1989learnability}.

We consider a finite set $\mathcal{F}$ of model $\model$ and suppose to have an algorithm able to find the model with the lowest empirical risk. We aim to bound the difference between the theoretical risk of the selected model and the theoretical risk of the better model in the set.
\\
We assume here that all models $\model$ are independent from the set of data use to compute the empirical risk, this hypothesis has been relaxed in previous work \cite{mohristabilitynodate}. However, in order to provide a simple example, we keep this hypothesis. 
\\
Formally, we define : $\tilde{f} = \underset{\hat{f} \in \mathcal{F}}{\text{argmin}} \left(\remp{\model} \right) \text{ and } f^* = \underset{\hat{f} \in \mathcal{F}}{\text{argmin}} \left( \rth{\model} \right)$ and we aim to bound $ \rth{\tilde{f}} - \rth{f^*}$.

\begin{cor}[Model selection on dependent non-causal field, tail inequality]
If $(\bf{H_1^\infty})$,$(\bf{H_2^\infty})$ and $(\bf{H_3})$ hold. For $\varepsilon >  \frac{4  n_{\Bar{B}} \mathbb{V}_{\infty} }{n} $,
    \begin{equation}
     \pr \left( \rth{\tilde{f}} - \rth{f^*} > \varepsilon \right) \leq 2 N \exp \left(\frac{ -2 n^2 \left(  \frac{\varepsilon}{2} - \frac{2  n_{\Bar{B}} \mathbb{V}_{\infty}}{n}  \right)^2  }{\left( n_{\Bar{B}} \mathbb{V}_{\infty} \right)^2 \left( 1 + A n_{\Bar{B}} n_B^3 \kappa!^2 \left\lceil \ln(n) \right\rceil^\kappa n \right)} \right).
     \label{prModelSelectStrong}
    \end{equation}
If $(\bf{H_1^m})$, $(\bf{H_2^m})$, $(\bf{H_3})$ and $(\bf{H_4})$, hold. For $\varepsilon \geq \frac{4 L_1(n)}{n^{1-\frac{2}{m}}}$,
    \begin{align}
    &\pr \left( \rth{\tilde{f}} - \rth{f^*} > \varepsilon \right) 
    \leq N \left( 2 \exp \left(  \frac{-  2 n^{2-\frac{4}{m}} \left( \frac{\varepsilon}{4} - \frac{L_1(n)}{n^{1-\frac{2}{m}}}  \right)^2}{ \left(H n_{\Bar{B}} \right)^2 \left(1 + E n_{\Bar{B}} n_B^3 (\kappa!)^2 \left\lceil \ln(n) \right\rceil^\kappa n \right) }\right) + \frac{\rho^m}{n} L_2(n) \right).
    \label{prModelSelectWeak}
    \end{align}
\end{cor}

\begin{proof}
\begin{align*}
     \rth{\tilde{f}} - \rth{f^*}  &=  \rth{\tilde{f}} - \remp{\tilde{f}} + \remp{\tilde{f}} - \remp{f^*} + \remp{f^*} - \rth{f^*}  \\
     &\leq \left( \rth{\tilde{f}} - \remp{\tilde{f}} \right) + \left( \remp{\tilde{f}} - \remp{f^*} \right) + \left( \remp{f^*} - \rth{f^*} \right) \\
     &\leq 2 \underset{\hat{f} \in \mathcal{F}}{\sup} | \rth{\hat{f}} - \remp{\hat{f}} | + \left( \remp{\tilde{f}} - \remp{f^*} \right) \\
     & \leq 2 \underset{\hat{f} \in \mathcal{F}}{\sup} | \rth{\hat{f}} - \remp{\hat{f}} | \quad \text{because $ \left( \remp{\tilde{f}} - \remp{f^*} \right) \leq 0$}.
\end{align*}

Thus, for all $\varepsilon$,
\begin{align*}
    \pr \left( \rth{\tilde{f}} - \rth{f^*} >\varepsilon \right) &\leq \pr \left(\underset{\hat{f} \in \mathcal{F}}{\sup} | \rth{\hat{f}} - \remp{\hat{f}} |> \frac{\varepsilon}{2} \right) \\
    & \leq \sum_{\model \in \mathcal{F}} \pr \left( | \rth{\model} - \remp{\model} | > \frac{\varepsilon}{2} \right) \quad \text{by union bound} \\
    & \leq N \pr \left( | \rth{\model} - \remp{\model} | > \frac{\varepsilon}{2} \right).
\end{align*}
Then, Theorem \ref{ErrGenTail} yields result for both hypotheses. 
\end{proof}

Bound for the expectation can also be obtained. We focus here on the case where hypotheses $(\bf{H_1^\infty})$, $(\bf{H_2^\infty})$ and $(\bf{H_3})$ hold. 

\begin{cor}[Model selection on dependent non-causal field, expectation inequality]
If $(\bf{H_1^\infty})$, $(\bf{H_2^\infty})$ and $(\bf{H_3})$ hold,
    \begin{equation*}
     \esp[\rth{\tilde{f}} - \rth{f^*}] \leq \left( \frac{\sqrt{\ln(N)}}{4} + \sqrt{\pi} \right) \sqrt{ \frac{2}{n} \left( \frac{B n_{\Bar{B}}^3}{n} + A n_{\Bar{B}}^3 n_{\mathcal{B}}^3 \kappa!^2  \ln(n)^\kappa \right)} + \frac{2C n_{\Bar{B}} }{\sqrt{n}}.
    \end{equation*}
    We can also provide an asymptotic equivalent,
    \begin{align*}
        \esp[\rth{\tilde{f}} - \rth{f^*}] \leq H(n_{\Bar{B}}, n_{\mathcal{B}}, \kappa, \rho, N, n) \underset{n \to \infty}{\sim} \left( \frac{\sqrt{\ln(N)}}{4} + \sqrt{\pi} \right) n_{\Bar{B}} n_{\mathcal{B}} \kappa! \sqrt{2An_{\Bar{B}} n_{\mathcal{B}}} \sqrt{\frac{{\ln(n)}^\kappa}{n}}.
    \end{align*}
    \label{ExpectModelSelectStrong}
\end{cor}

\begin{proof}
We showed in previous Corollary that $\rth{\tilde{f}} - \rth{f^*} \leq 2\underset{\hat{f} \in \mathcal{F}}{\sup} |\rth{\hat{f}} - \remp{\hat{f}}|$.
\\
Moreover, for all $s > 0$,
\begin{align*}
    \esp[\underset{\hat{f} \in \mathcal{F}}{\sup} |\rth{\hat{f}} - \remp{\hat{f}}|] &= \frac{1}{s} \esp \left[ \ln \left( \underset{\hat{f} \in \mathcal{F}}{\sup} \exp \left( s |\rth{\hat{f}} - \remp{\hat{f}}| \right) \right) \right] \\
    & \leq 
    \frac{1}{s} \ln \left( \esp \left[  \underset{\hat{f} \in \mathcal{F}}{\sup} \exp \left( s |\rth{\hat{f}} - \remp{\hat{f}}| \right)  \right] \right) \quad \text{using Jensen inequality} \\
    & \leq 
    \frac{1}{s} \ln \left( \esp \left[ \sum_{\hat{f} \in \mathcal{F}} \exp \left( s |\rth{\hat{f}} - \remp{\hat{f}}| \right)  \right] \right) \\
    & = 
    \frac{1}{s} \ln \left( N \esp \left[ \exp \left( s |\rth{\hat{f}} - \remp{\hat{f}}| \right)  \right] \right).
\end{align*} 

If $(\bf{H_1^\infty})$, $(\bf{H_2^\infty})$ and $(\bf{H_3})$ hold, applying Equation \eqref{strongexpMoment} from Corollary \ref{expMoment}. We get
\begin{align*}
\esp[\rth{\tilde{f}} - \rth{f^*}] 
\leq & \frac{2 \ln(N)}{s} + \frac{4 n_{\Bar{B}} \mathbb{V}_{\infty} }{n} + \frac{s \left( n_{\Bar{B}} \mathbb{V}_{\infty} \right)^2 \left( 1 + A n_{\Bar{B}} n_B^3 \kappa!^2 \left\lceil \ln(n) \right\rceil^\kappa n \right)}{4 n^2} \\ 
& + \frac{2 n_{\Bar{B}} \mathbb{V}_{\infty} }{n} \sqrt{2 \pi \left( 1 + A n_{\Bar{B}} n_B^3 \kappa!^2 \left\lceil \ln(n) \right\rceil^\kappa n \right)}, \; \text{using $\ln(1+x) \leq x$.}
\end{align*}
Then choosing $\displaystyle s = \frac{2n}{n_{\Bar{B}} \mathbb{V}_{\infty}} \sqrt{\frac{2 \ln(N)}{1 + A n_{\Bar{B}} n_B^3 \kappa!^2 \left\lceil \ln(n) \right\rceil^\kappa n}} $, we get 
\begin{align*}
    & \esp[\rth{\tilde{f}} - \rth{f^*}] \\
    &\leq \frac{2 n_{\Bar{B}} \mathbb{V}_{\infty}}{n} \left( \sqrt{\frac{\ln(N)}{2}\left( 1 + A n_{\Bar{B}} n_B^3 \kappa!^2 \left\lceil \ln(n) \right\rceil^\kappa n \right)} + 2 + \sqrt{2 \pi \left( 1 + A n_{\Bar{B}} n_B^3 \kappa!^2 \left\lceil \ln(n) \right\rceil^\kappa n \right)} \right) \\
    &\leq \frac{2 n_{\Bar{B}} \mathbb{V}_{\infty}}{n} \left(2 + \sqrt{\left( 1 + A n_{\Bar{B}} n_B^3 \kappa!^2 \left\lceil \ln(n) \right\rceil^\kappa n \right)} \left( \sqrt{\frac{\ln(N)}{2}} + \sqrt{2 \pi} \right) \right).
\end{align*}
\end{proof}

\begin{rem}
In the case where $\esp \left[ \exp \left( s |\rth{\hat{f}} - \remp{\hat{f}}| \right)  \right]$ does not exist, or if we only have  $(\bf{H_1^m})$, $(\bf{H_2^m})$, $(\bf{H_3})$ and $(\bf{H_4})$, it is possible to use a simpler bound.
It may be obtained noticing
\[\esp[\rth{\tilde{f}} - \rth{f^*}] \leq \esp[\underset{\hat{f} \in \mathcal{F}}{\sup} |\rth{\hat{f}} - \remp{\hat{f}}|] \leq N \esp[|\rth{\hat{f}} - \remp{\hat{f}}|]\] 
and using Corollaries \ref{simplStrongMomentIneq} or \ref{simplWeakMomentIneq}. If we do so, learning bound increase with $N$ instead of $\sqrt{\ln(N)}$. If we assume $(\bf{H_1^m})$, $(\bf{H_2^m})$, $(\bf{H_3})$ and $(\bf{H_4})$, we then get asymptotic bound  
\begin{equation*}
        \esp[\rth{\tilde{f}} - \rth{f^*}] \leq H(n_{\Bar{B}}, n_{\mathcal{B}}, \kappa, \rho, N, n) \underset{n \to \infty}{\sim} \frac{H \kappa! \sqrt{2E \pi (n_{\Bar{B}} n_B)^3 \ln(n)^\kappa}}{n^{\frac{1}{2}- \frac{2}{m}}}.
    \end{equation*}
\end{rem}
Where $H$ and $E$ are constants defined in Corollary \ref{simplWeakMomentIneq}.
\begin{rem}
In \cite{Lugosi2002}, a bound is provided for the expected maximal deviation in the i.i.d case,
\begin{equation*}
    \esp[\underset{\hat{f} \in \mathcal{F}}{\sup} |\rth{\hat{f}} - \remp{\hat{f}}|] \leq \sqrt{\frac{\ln(2N)}{2n}}.
\end{equation*}
Under our setting, with hypotheses $(\bf{H_1^\infty})$, $(\bf{H_2^\infty})$ and $(\bf{H_3})$, we obtained comparable bound (Corollary \ref{ExpectModelSelectStrong}) with an extra term $\sqrt{{\ln(n)}^\kappa}$.
\end{rem}

\section{Approximation}
\label{Sec:Concentration}
\subsection{Approximation of $X_t$}

\subsubsection{Notations}

Let's first recall and introduce some notations to handle random fields.

\begin{itemize}
    
    \item We already introduce the notation $\pmb{\varepsilon} = (\varepsilon_{t})_{t \in \mathbb{Z}^\kappa}$. Similarly, we use the notation $\pmb{\varepsilon}^\prime = (\varepsilon_{t}^\prime)_{t \in \mathbb{Z}^\kappa}$ where $\varepsilon_{t}^\prime$ are also random variables $\Omega \mapsto E$.
    \item For $ s \in \mathbb{Z}^\kappa$, $\theta_s$ will be the $s$-shift operator, ie for $ s \in \mathbb{Z}^\kappa, \theta_s(\inovField) = (\varepsilon_{s+t})_{t \in \mathbb{Z}^\kappa}$.
    \item For $ s \in \mathbb{Z}^\kappa$, we denote by $\pmb{\varepsilon}_s$ the $s$-shifted random field, ie $\pmb{\varepsilon}_s = \theta_s(\pmb{\varepsilon}) = \theta_s(\inovField) = (\varepsilon_{s+t})_{t \in \mathbb{Z}^\kappa}$.
\end{itemize}

\subsubsection{Exact reconstruction}

Theorem 1 from \cite{doukhan2007fixed} ensures, under suitable conditions on F, the existence and uniqueness of a function H such that, for each $t \in \mathbb{Z}^\kappa$

\begin{equation}
    X_t = H(\pmb{\varepsilon}_t).
\end{equation}

We make two comments on this result. 
\begin{itemize}
    \item This theorem provides an expression of each $X_t$ according to an infinite number of i.i.d. random variables (the whole random field $\pmb{\varepsilon}$). However many concentration inequalities involve only a finite number of random variable.
    \item This theorem relies on an absolute contraction hypothesis on F (similar to Equation \ref{stdContraction}) which is a stronger assumption than $(\bf{H_1^{\infty}})$ and $(\bf{H_1^{m}})$.
\end{itemize}
For those two reasons, we cannot use this \emph{exact} reconstruction of the solution $X_t$ of Equation \eqref{masterEquation}.

\label{Sec:Approximation}

\subsubsection{Intuition}

The idea is to approximate each $X_t$ by another random variable $\Tilde{X_t}$ which, similarly to $H(\pmb{\varepsilon})$, depends on the innovation $\pmb{\varepsilon}$.
\\
However, unlike $H(\pmb{\varepsilon})$, we will only use a finite number of random variable $\varepsilon_t$. The ones that are located in a finite neighborhood surrounding $X_t$.

\subsubsection{Definition}
\label{xiNotation}
Let's recall some notations and formally define the function $\hd{d}$.
\begin{itemize}
    \item In the equation $\forall t \in \mathbb{Z}^\kappa, X_t = F((X_{t+s})_{s \in \mathcal{B}}, \varepsilon_t)$ (Equation \eqref{masterEquation}), $\mathcal{B}$ is a $\kappa-$orthotope defined by $\mathcal{B} = \mathcal{V}(\delta,0) \backslash \{0\} = \prod_{i=1}^{\kappa} [-\delta_{i}, \delta_i] \backslash \{0\}$.
    \item We define the dilatation by $d$ of the $\kappa-$orthotope $\mathcal{V}(\delta,0)$ centered on $s$. With our notations, it corresponds to the $\kappa-$orthotope $\mathcal{V}(d\delta, s)$ defined for all $d$ in $\mathbb{N}$ and $s$ in $\mathbb{Z}^\kappa$ by 
    \begin{equation}
        \mathcal{V}(d\delta, s) = \{ t = (t_1, \dots, t_\kappa) \in (\mathbb{Z})^{\kappa} /  \, \forall i, s_i - d \delta_i \leq t_i \leq s_i + d \delta_i \}.
        \label{Vdelta}
    \end{equation}
    \item We define recursively the function $\hd{d}$ by
    \begin{equation}
        \hd{d}(X,  (\varepsilon_t)_{t \in \mathcal{V}(d \delta, t)}) = 
        \begin{cases}
            X \; \text{if d = 0}, \\
            F\left((\hd{d-1} \left(X, (\varepsilon_{u})_{u \in \mathcal{V}(\delta (d - 1), t+s)} \right) )_{s \in \mathcal{B}}, \varepsilon_t \right) \; \text{else}.
        \end{cases}
    \label{hd}
    \end{equation}
    \item Eventually, we are able to define the approximation. 
    \begin{equation}
    \forall d \in \mathbb{N}, \forall t \in \mathbb{Z}^\kappa, \Xt{t}{d} = \hd{d}(\Bar{X}, (\varepsilon_s)_{s \in \mathcal{V}(d \delta, t)}).
    \label{defApprox}
    \end{equation}
    Where $\Bar{X}$ is a random variable draw from the law $\mu_X$ independent from $(X_t)_{t \in \mathbb{Z}^\kappa}$ and $\pmb{\varepsilon}$.
\end{itemize}
We can reformulate Equation \eqref{hd} using the notation $\Xt{t}{d}$.

\begin{equation*}
    \forall d \in \mathbb{N}, \forall t \in \mathbb{Z}^\kappa, \Xt{t}{d} = 
    \begin{cases}
    \Bar{X} \; \text{if $d=0$}, \\
    F \left( (\Xt{t+s}{d-1})_{s \in \mathcal{B}}, \varepsilon_t \right) \; \text{else}.
    \end{cases}
\end{equation*}

In this way, $\Xt{t}{d}$ is an approximation of $X_t$ involving random variables $\varepsilon_t$ which belong to the finite neighborhood $\mathcal{V}(d\delta, t)$.
Outside, of this neighborhood, we complete  the approximation with a random variable $\Bar{X}$ draw form the law $\mu_X$ and independent from $(X_t)_{t \in \mathbb{Z}^\kappa}$ and $\pmb{\varepsilon}$.

\begin{rem}
    We emphasize that if the function H from Theorem 1 of \cite{doukhan2007fixed} exists and is unique, then $\displaystyle \lim_{d \to \infty} \hd{d}(X,  (\varepsilon_t)_{t \in \mathcal{V}(d \delta, t)}) = H(\pmb{\varepsilon})$.
    \\
    Nevertheless this limit might not exist and not be unique, but even in this case, for all finite $d$, the approximation $\hd{d}(X,  (\varepsilon_t)_{t \in \mathcal{V}(d \delta, t)})$ is always defined.  
\end{rem}

\subsubsection{Approximation error}
The approximation $\Xt{t}{d}$ is useful only if we are able to control the approximation error. That is the goal of the two following Lemma. 

\begin{lemma}
If $(\bf{H_1^m})$ and $(\bf{H_2^m})$ are verified, then
 \begin{equation*}
    \forall t \in \mathbb{Z}^\kappa, \forall d \in \mathbb{N}, \|X_t - \Xt{t}{d} \|_m \leq \rho^{d} \mathbb{V}_m.
 \end{equation*}
 \label{lem:approxF}
\end{lemma}
The proof can be found in the appendix \ref{proofApprox}.

Earlier, we define the statistic $\SI$ (see Equation \eqref{eq:SI_model}), which depends on random variables $(X_t)_{t \in \mathcal{I}}$. We now introduce an approximation of the statistic $\SI$ that we call $\SItilde$ relying on approximations $\Xt{t}{d}$ for $t$ in $\mathcal{I}$.

We recall that
\begin{equation}
    \SI = \sum_{t  \in \mathcal{I}} \Phi((X_{t+s})_{s \in \Bar{\mathcal{B}}}).
    \label{equ:SI}
\end{equation}

Similarly, we define
\begin{equation}
    \SItilde = \sum_{t  \in \mathcal{I}} \Phi((\Xt{t+s}{d})_{s \in \Bar{\mathcal{B}}}). \label{equ:SITilde}
\end{equation}

Using Lemma \ref{lem:approxF}, we are able to control the difference between $\SI$ and $\SItilde$. This is the purpose of the following corollary. 

\begin{cor}[Moment inequality for the approximation error]
We  assume $(\bf{H_1^m})$, $(\bf{H_2^m})$ and $(\bf{H_3})$. We recall that we denote $\card{I}$ by $n$ and $\card{\Bar{B}}$ by $n_{\Bar{B}}$. Then it holds 
\begin{equation}
     \| \SI - \SItilde \|_m \leq n n_{\Bar{B}} \rho^{d} \mathbb{V}_m.
     \label{eqApprox}
\end{equation}
If $(\bf{H_1^\infty})$ and  $(\bf{H_2^{\infty}})$ are verified, $\| \SI - \SItilde \|_\infty = \sup|\SI - \SItilde|$ and it holds almost surely
\begin{equation} 
|\SI - \SItilde| \leq n n_{\Bar{B}} \rho^{d} \mathbb{V}_{\infty}.
\end{equation}
\label{cor:approx}
\end{cor}

\begin{proof}

By definition of $\SI$ and $\SItilde$ (Equations \eqref{equ:SI} and \eqref{equ:SITilde}), using $(\bf{H_3})$, it holds

\begin{equation*}
    \| \SI - \SItilde \|_m \leq \sum_{t \in \mathcal{I}} \| \Phi((X_{t+s})_{s \in \Bar{\mathcal{B}}}) - \Phi((\Xt{t+s}{d})_{s \in \Bar{\mathcal{B}}})\|_m
    \leq  \sum_{t  \in \mathcal{I}} \sum_{s \in \Bar{\mathcal{B}}} \| X_{t+s} - \Xt{t+s}{d}\|_m.
\end{equation*}
Then, using Lemma \ref{lem:approxF}
\begin{equation*}
    \| \SI - \SItilde \|_m \leq n n_{\Bar{B}} \rho^{d} \mathbb{V}_m.
\end{equation*}

\end{proof}

\subsection{Concentration inequality for $\SItilde$}

In this subsection, we will establish a concentration inequality for $\SItilde$. Indeed, $\SItilde$ may be seen as a function of a finite number of random variables, thus it is possible to establish a McDiarmid's inequality for $\SItilde$.

More precisely, we aim to show the following theorem.

\begin{theorem}[Tail inequality for the approximation]
Let $d \in \mathbb{N}$.
If we assume $(\bf{H_1^\infty})$, $(\bf{H_2^{\infty}})$ and $(\bf{H_3})$
\begin{equation}
     \forall \varepsilon >0, \mathbb{P} \left( |\SItilde - \esp \left[ \SItilde \right]  | \geq \varepsilon \right) \leq 2 \exp \left( \frac{-2\varepsilon^2 }{ \left(n_{\Bar{B}} \mathbb{V}_{\infty} \right)^2 \left( n^2 \rho^{2d} + N_2 \left( n_{B} \Upsilon(d) \right)^2 \right)} \right).
\end{equation}
If we assume $(\bf{H_1^m})$, $(\bf{H_2^m})$, $(\bf{H_2})$ and $(\bf{H_3})$, $\forall (t_1, t_2) > 0,$
\begin{equation}
    \forall \varepsilon > 2npM + p\Bar{c}, \, \pr \left( |\SItilde - \esp \left[ \SItilde \right]  | \geq \varepsilon \right) \leq 2 \left( p + \exp \left( \frac{-2 (\varepsilon - (p \Bar{c} + 2npM ) )^2}{t_1^2 + N_2 t_2^2} \right) \right),
\end{equation}
with
\begin{equation*}
    p \leq  \left( \frac{ n n_{\Bar{B}} \rho^{d} \mathbb{V}_{m}}{t_1} \right)^m + N_2 \left( \frac{n_{\Bar{B}}n_{B} \mathbb{V}_m \Upsilon(d) }{t_2} \right)^m \; \text{and} \; \Bar{c} = t_1 + N_2 t_2,
\end{equation*}
and
\begin{equation*}
    \Upsilon(d) \; \text{such that} \; \Upsilon(d) \leq \upsilon = \frac{1}{n_B} + \frac{\kappa!}{\ln(\rho^{-1})^{\kappa}} + \kappa \left( \frac{\kappa-1}{\ln(\rho^{-1})e} \right)^{\kappa-1} \underset{\kappa \to \infty}{\sim} \frac{\kappa!}{\ln(\rho^{-1})^{\kappa}}.
\end{equation*}
\label{theo:dev_bound}
\end{theorem}

To that end, we have to fulfill the required assumptions for McDiarmid's inequality. First, we need to recall and add some notations.

\subsubsection{Notations}
\label{notations}
\begin{itemize}
    \item $n = \card{\mathcal{I}}$, the number of variables $X_t$ appearing in $\SI$ (see Section \ref{statSI}).
    \item We recall that $\mathcal{V}(\delta,0) = \prod_{i=1}^{\kappa} [-\delta_{i}, \delta_i]$, then $\mathcal{B} = \mathcal{V}(\delta,0) \backslash \{0\}$ and $\Bar{\mathcal{B}} = \mathcal{V}(\Bar{\delta},0) $.
    \\
    Additionally, we introduce notations $\mathcal{B} + s$ and $\Bar{\mathcal{B}} + s$ which denote the set $\mathcal{B}$ (resp. $\Bar{\mathcal{B}}$) shifted by $s$, i.e., $\mathcal{B} + s = \mathcal{V}(\delta,s) \backslash \{ s \}$ (resp. $\Bar{\mathcal{B}} + s = \mathcal{V}(\Bar{\delta},s) $).
    \item $n_B = \card{\mathcal{B} \bigcup \{0\}} = \card{\mathcal{B}} +1$, the number of random variables in the $\kappa-$orthotope $\mathcal{B}$ plus the point $\{0\}$.
    \item $n_{\Bar{B}} =  \card{ \Bar{\mathcal{B}}}$ the number of components of $\Phi$ (see Section \ref{statSI}).
    \item $\displaystyle n_d = \card{ \mathcal{V}(d \delta, t)  }$ the number of random marginal variables $\varepsilon_t$ of $\pmb{\varepsilon}$ used in the approximation $\Xt{t}{d}$ (see Equation \ref{defApprox}). 
    \item $\displaystyle N_1 =  \card{\bigcup_{t \in \mathcal{I}} \left( \mathcal{V}(\Bar{\delta}, t) \right) } = \card{ \bigcup_{t \in \mathcal{I}} \left( \Bar{\mathcal{B}} + t \right) }$ the number of random variables $\Xt{t}{d}$ occurring in $\SItilde$.
    \item $\displaystyle N_2 = \card{\bigcup_{t \in \mathcal{I}} \left( \bigcup_{s \in \Bar{\mathcal{B}} + t} \left( \mathcal{V}(d\Bar{\delta}, t)  \right) \right)}$
    number of random variables $\varepsilon_t$ occurring in $\SItilde$.
\end{itemize}

There are some relations between these numbers.
\begin{lemma}
~\begin{itemize}
    \item $n \leq N_1 \leq n n_{\Bar{B}}$ \quad \text{and} \quad $N_1 \leq N_2 \leq N_1 n_d $. 
    \item $\displaystyle n_d = \card{\mathcal{V}(d \delta, t)} = \prod_{i=1}^\kappa \left( 2d\delta_i + 1 \right) \leq d^\kappa \prod_{i=1}^\kappa \left( 2\delta_i + \frac{1}{d} \right) \leq d^\kappa n_{B}$.
\end{itemize}
\label{notation}
\end{lemma}

Higher bounds for $N_1$ and $N_2$ are reached when above unions involve pairwise disjoint sets. Nevertheless, it is not often the case in practice. For instance, in machine learning settings, the training and validation sets are usually connected spaces. Therefore, adding more hypotheses on the topology of $\mathcal{I}$ should improves these bounds.

\subsubsection{An extention of McDiarmid's inequality}

In order to obtain a concentration inequality for $\SItilde$, we need a specific McDiarmid's inequality. Indeed, when we only assume the weaker hypotheses $(\bf{H_1^m})$ and $(\bf{H_2^m})$ the uniform difference-bound hypothesis (as defined as in \cite{kutin2002extensions}) is not fulfilled, thus classical McDiarmid's inequality \cite{mcdiarmid1989} don't hold. 
\\
That's why, we need an extended version of McDiarmid's inequality which holds even when the bounded difference hypothesis is verified only with high probability. Several results of this type exist (\cite{kutin2002extensions,combesExtension2015,pmlr-v32-kontorovicha14,warnke_2016}). Here, we chose to use the extended McDiarmid's inequality from \cite{combesExtension2015}.
\\
We stand below the "$A$-difference bounded" assumption which correspond to Assumption 1.2. from \cite{combesExtension2015} and the extension of the McDiarmid's inequality (Theorem 2.1. from\cite{combesExtension2015}).

\begin{defini}[$A$-difference-bound]
\label{def:pdifferencebound}
$f : \prod_{i=1}^N \Omega_i  \to \mathbb{R} $ is $p$-difference bounded by $(c_j)_{j \in [1,N]}$ if and only if.
\begin{itemize}
    \item It exists $A \subset \prod_{i=1}^N \Omega_i$.
    \item For $(w,w^\prime) \in  A^2$ such that $w$ and $w^\prime$ differ only in the j-th coordinate ,$|f(w) - f(w^\prime)| \leq c_j$.
\end{itemize}
\end{defini}

\begin{lemma}[$p$-difference-bound McDiarmid, Theorem 2.1. from \cite{combesExtension2015}]
If f is $\mathbb{P}(A)$-difference bounded by $(c_j)_{j \in [1,N]}$.
\begin{equation*}
    \forall t > 0, \pr(|f(X_1, \dots, X_n) - \esp\left[ f(X_1, \dots, X_n) | A \right]| \ge t) \leq 2 \left( p +  \exp \left( \frac{-2(t - p \Bar{c})^2}{\sum_{j=1}^N c_j^2} \right) \right),
\end{equation*}
with $\Bar{c} = \sum_{j=1}^N c_i$ and $p = 1 - \mathbb{P}(A)$.
\label{lem:extendedMacDiarmid}
\end{lemma}

\subsubsection{Difference bound for $\SItilde$}

To verify those assumptions (strong difference bound or A-difference bound), we need to bound the difference between the statistic $\SI$ and the statistic $\SI$ when one of its composing random variables is replaced by an i.i.d copy.
To handle such case, we need to introduce further notations.

We recall that
\begin{equation*}
    \SItilde = \sum_{t \in \mathcal{I}} \Phi((\Xt{t+s}{d})_{s \in \Bar{\mathcal{B}}}) = \sum_{t  \in \mathcal{I}} \Phi(\hd{d}(\Bar{X}, (\varepsilon_s)_{s \in \mathcal{V}(d \delta, t)})).
\end{equation*}

Two types of random variables are involved in $\SItilde$.
\begin{itemize}
    \item The marginal variables $\varepsilon_s$ (for $\displaystyle s \in \bigcup_{t \in \mathcal{I}} \mathcal{V}(d\Bar{\delta},t)$).
    \item The (unique) variable $\Bar{X}$. 
\end{itemize}

We introduce the following notations.

\begin{itemize}
    \item For all $t$ in $\mathbb{Z}^\kappa$, $\Xtt{t}{d} = \hd{d}(\Bar{X}^\prime, (\varepsilon_s)_{s \in \mathcal{V}(d \delta, t)})$ where $\Bar{X}^\prime$ is drawn from the law $\mu_X$ and independent from $(X_t)_{t \in \mathbb{Z}^\kappa}$, $\pmb{\varepsilon}$ and $\Bar{X}$.
    
    \item For all $t$ in $\mathbb{Z}^\kappa$ and for all $i$ in $\mathcal{V}(d\Bar{\delta},t)$, $\Xt{t}{d,i} = \hd{d}(\Bar{X}, (\varepsilon_s)_{s \in \mathcal{V}(d \delta, t) \backslash \{i\} } \bigcup \varepsilon_i^\prime )$ where $\varepsilon_i^\prime$ is drawn from the law $\mu_\varepsilon$ and independent from $(X_t)_{t \in \mathbb{Z}^\kappa}$, $\pmb{\varepsilon}$ and $\Bar{X}$.
    \\
    For all $i$ in $\mathbb{Z}^\kappa$ not in $\mathcal{V}(d\Bar{\delta},t)$, we define $\Xt{t}{d,i} = \Xt{t}{d}$.
    
    \item For all $t$ in $\mathbb{Z}^\kappa$, $\displaystyle \SItildep = \sum_{t \in \mathcal{I}} \Phi((\Xtt{t+s}{d})_{s \in \Bar{\mathcal{B}}})$.
    
    \item For all $t$ in $\mathbb{Z}^\kappa$ and for all $i$ in $\displaystyle \bigcup_{t \in \mathcal{I}} \mathcal{V}(d\Bar{\delta},t)$, $\displaystyle \SItildei = \sum_{t \in \mathcal{I}} \Phi(\Xt{t}{d,i})$.
\end{itemize}

\begin{lemma}
If we assume $(\bf{H_1^m})$ and $(\bf{H_2^m})$,
\begin{itemize}
    \item For all $t$ in $\mathbb{Z}^\kappa$, $\| \Xt{t}{d} - \Xtt{t}{d} \|_m \leq \rho^d \mathbb{V}_m$.
    \item For all $t$ and $i$ in $\mathbb{Z}^\kappa$.
    \begin{itemize}
        \item If $i \notin \mathcal{V}(d\Bar{\delta},t)$, $\Xt{t}{d} = \Xt{t}{d,i}$.
        \item Else, there is a unique $c \in [0, d]$ such that $i \in \mathcal{V}(c\delta, t)$ and $i \notin \mathcal{V}((c-1)d\delta, t)$ and  $\| \Xt{t}{d} - \Xt{t}{d,i} \|_m \leq \rho^c \mathbb{V}_m$.
    \end{itemize}
\end{itemize}
If we assume $(\bf{H_1^\infty})$ and $(\bf{H_2^\infty})$, results are similar except that they hold with the norm $\|.\|_\infty$.
\label{bdiXt}
\end{lemma}

\begin{proof}
We assume $(\bf{H_1^m})$ and $(\bf{H_2^m})$.
\begin{itemize}
    \item The demonstration of the first point is the same as Lemma \ref{lem:approxF}.
    \item Firstly, if $i \notin \mathcal{V}(d\delta, t)$, by definition of $\Xt{t}{d,i}$, $\Xt{t}{d,i} = \Xt{t}{d}$.
    \\
    Secondly, if $i \in \mathcal{V}(d\delta, t)$, we can write $\displaystyle \mathcal{V}(d\delta, t) = \bigcup_{c=0}^d \mathcal{V}(c\delta, t) \backslash \mathcal{V}((c-1)\delta, t)$ with $\mathcal{V}(0, t) = \{t\}$ and $\mathcal{V}(-1, t) = \emptyset$.
    \\
    Moreover, for all $c_1 \ne c_2$, $\displaystyle \mathcal{V}(c_1\delta, t) \backslash \mathcal{V}((c_1-1)\delta, t) \bigcap \mathcal{V}(c_2\delta, t) \backslash \mathcal{V}((c_2-1)\delta, t) = \emptyset$.
    \\
    Consequently, it exists an unique $c$ in $[0,d]$ such that $i \in \mathcal{V}(c\delta, t)$ and $i \notin \mathcal{V}((c-1)\delta, t)$. Then with the same argument as Lemma \ref{lem:approxF}, we can show that $\| \Xt{t}{d} - \Xt{t}{d,i} \|_m \leq \rho^c \mathbb{V}_m$.
\end{itemize}
\end{proof}

We can now bound the difference between $\SItilde$ and $\SItildep$ and $\SItilde$ and $\SItildei$.

\begin{lemma}[Difference bound for $\SItilde$]
~\begin{itemize}
\item If we assume $(\bf{H_1^m})$, $(\bf{H_2^m})$ and $(\bf{H_3})$, it holds
\begin{equation*}
    \| \SItilde - \SItildep \|_m \leq n n_{\Bar{B}} \rho^{d} \mathbb{V}_m.
\end{equation*}
And,
\begin{equation*}
    \|\SItilde - \SItildei \|_m \leq n_{\Bar{B}} \mathbb{V}_m \left(1 + n_{\mathcal{B}} \kappa \sum_{c=1}^d c^{\kappa - 1} \rho^c \right).
\end{equation*}

\item If we assume $(\bf{H_1^\infty})$, $(\bf{H_2^{\infty}})$ and $(\bf{H_3})$, it holds almost surely
\begin{equation*}
    | \SItilde - \SItildep | \leq n n_{\Bar{B}} \rho^{d} \mathbb{V}_\infty.
\end{equation*}
And,
\begin{equation*}
    | \SItilde - \SItildei | \leq n_{\Bar{B}} \mathbb{V}_\infty \left(1 + n_{\mathcal{B}} \kappa \sum_{c=1}^d c^{\kappa - 1} \rho^c \right).
\end{equation*}

\end{itemize}
\label{momentbdi}
\end{lemma}

The proof of this lemma can be found in the appendix \ref{proveBDI}. 

In previous Lemma, the quantity $\sum_{c=1}^d c^{\kappa - 1} \rho^c$ occurs, we show in the next Lemma that we can bound this quantity independently of d.

\begin{lemma}
If $(\bf{H_1^m})$, $(\bf{H_2^m})$, and $(\bf{H_3})$. For all $d \in \mathbb{N}$, it holds
    \begin{equation*}
        \| \SItildei - \SItilde \|_m \leq n_{\Bar{B}}n_{B} \mathbb{V}_m \Upsilon(d).
    \end{equation*}
The function $\Upsilon$ is defined by
\begin{equation*}
    \Upsilon(d) =
    \left\{
    \begin{array}{l}
        \displaystyle
        \kappa \left( \frac{1}{n_B \kappa} + \frac{(\kappa -1)!}{\ln(\rho^{-1})^{\kappa}} \left( 1 - \rho^{d+1} \sum_{i=0}^{\kappa -1} \frac{\left( (d+1) \ln(\rho^{-1}) \right)^i}{i!} \right) \right), \; \text{ if } d < \left\lfloor \frac{\kappa -1}{\ln(\rho^{-1})} \right\rfloor. \\
        \displaystyle
        \kappa \left( \frac{1}{n_B \kappa} + \frac{(\kappa -1)!}{\ln(\rho^{-1})^{\kappa}} \left( 1 - e^{- (\kappa - 1)} \sum_{i=0}^{\kappa -1} \frac{\left( \left\lfloor\frac{\kappa - 1}{\ln(\rho^{-1})} \right\rfloor \ln(\rho^{-1}) \right)^i}{i!} \right) + \left( \frac{\kappa-1}{\ln(\rho^{-1})e} \right)^{\kappa-1} \right), \; \text{ if } d = \left\lfloor \frac{\kappa -1}{\ln(\rho^{-1})} \right\rfloor. \\
        \displaystyle
        \kappa \left( \frac{1}{n_B \kappa} + \frac{(\kappa -1)!}{\ln(\rho^{-1})^{\kappa}} \left( 1 - \rho^d \sum_{i=0}^{\kappa - 1} \frac{\left( d \ln(\rho^{-1}) \right)^i}{i!} \right) + \left( \frac{\kappa-1}{\ln(\rho{-1})e} \right)^{\kappa-1} \right), \; \text{ if } d > \left\lfloor \frac{\kappa -1}{\ln(\rho^{-1})} \right\rfloor.
    \end{array}
    \right.
\end{equation*}
With
\begin{equation*}
     \Upsilon(d) \leq \upsilon = \frac{1}{n_B} + \frac{\kappa!}{\ln(\rho^{-1})^{\kappa}} + \kappa \left( \frac{\kappa-1}{\ln(\rho{^{-1}})e} \right)^{\kappa-1}.
\end{equation*}
We point out that the constant $\upsilon$ is independent from $d$ and $\upsilon \underset{\kappa \to \infty}{\sim} \frac{\kappa!}{\ln(\rho^{-1})^\kappa}$ (using Stirling formula).
\label{lem:technical}
\end{lemma}

The proof of this lemma can be found in the appendix \ref{techniLemma}.

\subsubsection{McDiarmid's inequality}

We can now use McDiarmid's inequality for $\SItilde$ and prove Theorem \ref{theo:dev_bound}. We distinguish two cases.

\begin{itemize}
    \item If $(\bf{H_1^\infty})$ and $(\bf{H_1^\infty})$ are verified, the standard bounded difference hypothesis (Equation \eqref{equ:strong_difference_bound}) is fulfilled. Therefore, we can directly apply the classical McDiarmid's inequality for i.i.d. random variables (\cite{mcdiarmid1989}). This is the purpose of Lemma \ref{uniforDev}.
    \item If $(\bf{H_1^m})$ and $(\bf{H_1^m})$ are verified for only a finite $m \in \mathbb{N}$, the standard bounded difference hypothesis is not fulfilled. Thus we will use the notion of "$A$-difference bound" and apply Lemma \ref{lem:extendedMacDiarmid}. This is the point of Lemma \ref{mDev}.
\end{itemize}

\begin{lemma}
If $(\bf{H_1^\infty})$,$(\bf{H_2^\infty})$ and $(\bf{H_3})$. Let $d \in \mathbb{N}$. It holds
\begin{align*}
    \pr \left( |\SItilde - \esp \left[ \SItilde \right]  | \geq \varepsilon \right) &\leq 2 \exp \left(- 2 \frac{\varepsilon^2}{(n n_{\Bar{B}} \rho^{d} \mathbb{V}_{\infty})^2 + \sum\limits_{i=1}^{N_2} \left( n_{\Bar{B}} n_{B} \mathbb{V}_\infty \Upsilon(d) \right)^2 } \right) \\
    & \leq 2 \exp \left( \frac{-2\varepsilon^2 }{ \left(n_{\Bar{B}} \mathbb{V}_{\infty} \right)^2 \left( n^2 \rho^{2d} + N_2 \left( n_{B} \Upsilon(d) \right)^2 \right)} \right).
\end{align*}
\label{uniforDev}
\end{lemma}
\begin{proof}
Using Lemma \ref{momentbdi} and \ref{lem:technical} with hypotheses $(\bf{H_1^\infty})$ and $(\bf{H_2^\infty})$, we get
\begin{itemize}
    \item $\displaystyle | \SItilde - \SItildep | \leq c_i \; \text{with} \; c_i = n n_{\Bar{B}} \rho^{d} \mathbb{V}_{\infty}.$
    \item $\displaystyle \forall i \in \bigcup_{t \in \mathcal{I}} \mathcal{V}(d\Bar{\delta},t),  | \SItilde - \SItildei | \leq c_i \; \text{with} \; c_i = n_{\Bar{B}} n_{B} \mathbb{V}_\infty \Upsilon(d).$
\end{itemize}
Applying McDiarmid's inequality yields the result. 
\end{proof}

\begin{lemma}
If we assume $(\bf{H_1^m})$, $(\bf{H_2^m})$, $(\bf{H_3})$ and $(\bf{H_4})$. Let $d \in \mathbb{N}$, and $t_1, t_2 >0$. There is $p \in (0,1)$, such that, for $\varepsilon \geq 2npM + p\Bar{c}$.
\begin{equation*}
\pr \left( |\SItilde - \esp \left[ \SItilde \right]  | \geq \varepsilon \right) \leq 2 \left( p + \exp \left( \frac{-2 (\varepsilon - (p \Bar{c} + 2npM ) )^2}{t_1^2 + N_2 t_2^2} \right) \right),
\end{equation*}
with
\begin{equation*}
    p \leq  \left( \frac{ n n_{\Bar{B}} \rho^{d} \mathbb{V}_{m}}{t_1} \right)^m + N_2 \left( \frac{n_{\Bar{B}}n_{B} \mathbb{V}_m \Upsilon(d) }{t_2} \right)^m \; \text{and} \; \Bar{c} = t_1 + N_2 t_2.
\end{equation*}
\label{mDev} 
\end{lemma}

\begin{proof}
First, we establish a $A$-difference bound (Definition \ref{def:pdifferencebound}).
Let $t_1, t_2 > 0$, using Lemma \ref{momentbdi}, \ref{lem:technical} and Markov's inequality, it holds
\begin{equation*}
    \forall t_1 > 0, \pr(|\SItilde -  \SItildep | \geq t_1) \leq \left( \frac{ n n_{\Bar{B}} \rho^{d} \mathbb{V}_{m}}{t_1} \right)^m.
\end{equation*}
    Then, for all $\displaystyle i \in \bigcup_{t \in \mathcal{I}} \mathcal{V}(d\Bar{\delta},t),
    \; \text{for all} \; t_2 > 0, \pr(|\SItilde -  \SItildei | \geq t_2) \leq \left( \frac{n_{\Bar{B}}n_{B} \mathbb{V}_m \Upsilon(d) }{t_2} \right)^m.$

We define the event $\displaystyle A = \bigcup_{i \in \bigcup_{t \in \mathcal{I}} \mathcal{V}(d\Bar{\delta},t)} \left( |\SItilde -  \SItildei | > t_2 \right) \bigcup \left( |\SItilde -  \SItildep | > t_1 \right)$ and $p= 1 - \pr(A)$. \\
By union bound, it holds
    $\displaystyle p \leq  \left( \frac{ n n_{\Bar{B}} \rho^{d} \mathbb{V}_{m}}{t_1} \right)^m + N_2 \left( \frac{n_{\Bar{B}}n_{B} \mathbb{V}_m \Upsilon(d) }{t_2} \right)^m.$
We also define $\Bar{c} = t_1 + N_2 t_2$. Therefore, applying Theorem 2.1 from \cite{combesExtension2015}, we get
\begin{equation*}
    \forall \varepsilon > 0, \pr \left( |\SItilde - \esp \left[ \SItilde | A \right]  | \geq \varepsilon \right) \leq 2 \left( p + \exp \left( \frac{-2 (\varepsilon - p \Bar{c} )^2}{t_1^2 + N_2 t_2^2} \right) \right).
\end{equation*}
Moreover
\begin{equation*}
    \esp \left[ \SItilde \right] = \pr(A) \esp \left[ \SItilde | A \right] + \pr(\Bar{A}) \esp \left[ \SItilde | \Bar{A} \right].
\end{equation*}
Then
\begin{align*}
    |\esp \left[ \SItilde \right] - \esp \left[ \SItilde | A \right]|  &= \left( 1 - \pr(A) \right) |\esp \left[ \SItilde | \Bar{A} \right] - \esp \left[ \SItilde | A \right]| \\
    & \leq p 2 n M \; \text{with} \;  M = \| \Phi \|_\infty \; (\text{see Hypothesis} \, (\bf{H_4})).
\end{align*}
Eventually, we get
\begin{align*}
    \forall \varepsilon > 2npM, \pr \left( |\SItilde - \esp \left[ \SItilde \right]  | \geq \varepsilon \right) & = \pr \left( |\SItilde - \esp \left[ \SItilde | A \right] + \esp \left[ \SItilde | A \right]  - \esp \left[ \SItilde \right]  | \geq \varepsilon \right) \\
    & \leq \pr \left( |\SItilde - \esp \left[ \SItilde | A \right] | \geq \varepsilon - 2npM \right) \\
    & \leq 2 \left( p + \exp \left( \frac{-2 (\varepsilon - (p \Bar{c} + 2npM ) )^2}{t_1^2 + N_2 t_2^2} \right) \right).
\end{align*}
\end{proof}

\section{Concentration inequalities for $\SI$}
\label{Sec:Deviation}
\subsection{General concentration inequalities}

Our initial goal is to provide concentration inequalities for $\SI$. It can be achieved using concentration inequalities from previous section (involving $\SItilde$) and Lemma \ref{lem:approxF}. 

\begin{lemma}[General concentration inequality for $\SI$]
Let $d \in \mathbb{N}$. If $(\bf{H_1^\infty})$, $(\bf{H_2^\infty})$ and $(\bf{H_3})$ are verified, the following relation holds for $\varepsilon > 2 n n_{\Bar{B}} \rho^{d} \mathbb{V}_{\infty}$.
\begin{equation}
 \pr \left( |\SI - \esp \left[ \SI \right]  | \geq \varepsilon \right) \leq 2 \exp \left( \frac{-2\left(\varepsilon - 2n n_{\Bar{B}} \rho^{d} \mathbb{V}_{\infty} \right)^2 }{ \left(n_{\Bar{B}}\mathbb{V}_{\infty} \right)^2 \left( n^2 \rho^{2d} + N_2 \left( n_{B} \Upsilon(d) \right)^2 \right)} \right).
     \label{equ:concentrationSI_1}
\end{equation}
If $(\bf{H_1^m})$, $(\bf{H_2^m})$, $(\bf{H_3})$ and $(\bf{H_4})$, are verified. For $t_1, t_2 > 0$, there is $p \in (0,1)$ such that for $ \varepsilon > 4npM + 2p\Bar{c}$.
\begin{equation}
    \pr \left( |\SI - \esp \left[ \SI \right]  | \geq \varepsilon \right) \leq 2 \left( p + \exp \left( \frac{-2 (\frac{\varepsilon}{2} - (p \Bar{c} + 2npM ) )^2}{t_1^2 + N_2 t_2^2} \right) \right) + \left( \frac{2n n_{\Bar{B}} \rho^{d} \mathbb{V}_{m}}{\varepsilon} \right)^m,
    \label{equ:concentrationSI_2}
\end{equation}
with
\begin{equation*}
    p \leq  \left( \frac{ n n_{\Bar{B}} \rho^{d} \mathbb{V}_{m}}{t_1} \right)^m + N_2 \left( \frac{n_{\Bar{B}}n_{B} \mathbb{V}_m \Upsilon(d) }{t_2} \right)^m \; \text{and} \; \Bar{c} = t_1 + N_2 t_2.
\end{equation*}
\label{concentrationSI}
\end{lemma}

\begin{proof}
~\paragraph{If $(\bf{H_1^\infty})$ and $(\bf{H_3})$ are verified.}
Using Corollary \ref{cor:approx}
\begin{align*}
|\SI -\esp[\SI]| &= | \SI - \SItilde + \SItilde - \esp\left[ \SItilde \right] + \esp\left[ \SItilde \right] - \esp\left[ \SI \right] | \\
&\leq | \SI - \SItilde | +  |\SItilde -\esp[\SItilde]| + | \esp\left[ \SItilde \right] - \esp\left[ \SI \right]|  \\
&\leq |\SItilde -\esp[\SItilde]| + 2n n_{\Bar{B}} \rho^{d} \mathbb{V}_{\infty}.
\end{align*}
Therefore, for $ \varepsilon > 2 n n_{\Bar{B}} \rho^{d} \mathbb{V}_{\infty}$.
\begin{align*}
 \pr \left( |\SI - \esp \left[ \SI \right]  | \geq \varepsilon \right) &\leq \pr \left( |\SItilde - \esp \left[ \SItilde \right]  | \geq \varepsilon - 2 n n_{\Bar{B}} \rho^{d} \mathbb{V}_{\infty} \right) \\
     & \leq 2 \exp \left( \frac{-2\left(\varepsilon - 2n n_{\Bar{B}} \rho^{d} \mathbb{V}_{\infty} \right)^2 }{ \left(n_{\Bar{B}}\mathbb{V}_{\infty} \right)^2 \left( n^2 \rho^{2d} + N_2 \left( n_{B} \Upsilon(d) \right)^2 \right)} \right).
\end{align*}

\paragraph{If $(\bf{H_1^m})$, $(\bf{H_2^m})$ and $(\bf{H_3})$ are verified.} 
Using Corollary \ref{cor:approx} and Markov inequality,
\begin{equation*}
    \forall t > 0, \pr \left( | \SI - \SItilde | \geq t \right) \leq \left( \frac{n n_{\Bar{B}} \rho^{d} \mathbb{V}_{m}}{t} \right)^m.
\end{equation*}
Then
\begin{align*}
    \pr  \left(| \SI - \esp[\SI] | \geq \varepsilon \right) &= \pr \left( | \SI - \SItilde + \SItilde - \esp[\SI] | \geq \varepsilon \right) \leq \pr \left( | \SI - \SItilde | + | \SItilde - \esp[\SI] | \geq \varepsilon \right) \\
    &\leq \pr  \left(| \SI - \SItilde | \geq \frac{\varepsilon }{2} \right) + \pr  \left(| \SItilde - \esp[\SI] | \geq \frac{\varepsilon }{2} \right) \\
    & \leq 2 \left( p + \exp \left( \frac{-2 (\frac{\varepsilon}{2} - (p \Bar{c} + 2npM ) )^2}{t_1^2 + N_2 t_2^2} \right) \right) + \left( \frac{2n n_{\Bar{B}} \rho^{d} \mathbb{V}_{m}}{\varepsilon} \right)^m.
\end{align*}
\end{proof}

\begin{rem}
The choice of the parameter d is a trade off between the quality of the approximation and the number of random variables involved in the McDiarmid's inequality. Indeed, When $d$ is high, the approximation error decreases (see Lemma \ref{momentbdi}), conversely the number of random variables $N_2$ increases (see the bound for $N_2$ in the Lemma \ref{notation}).
\end{rem}

\subsection{Concentration inequalities with optimized parameters}

\label{fullResults}

Equations \eqref{equ:concentrationSI_1} and \eqref{equ:concentrationSI_2} in Lemma \ref{concentrationSI} has been established for any $d \in \mathbb{N}$ and any positive value of $t_1, t_2$. Therefore, we can choose a suitable value for each of these parameters to improve our bounds.
\\
Moreover, the quantity $N_2$ is not easy to interpret and even more to estimate. Thus, in the following theorems, we fix parameters $d, t_1, t_2$ and replace $N_2$ by an upper bound corresponding to the worst case.

\begin{theorem}[Improved concentration inequality for $\SI$, uniform contraction case]
    If $(\bf{H_1^\infty})$, $(\bf{H_2^\infty})$ and $(\bf{H_3})$, for $\varepsilon > 2 n_{\Bar{B}} \mathbb{V}_{\infty}$
    \begin{align*}
     \pr \left( |\SI - \esp \left[ \SI \right]  | \geq \varepsilon \right) \leq 2 \exp \left( \frac{-2\left(\varepsilon - 2 n_{\Bar{B}} \mathbb{V}_{\infty} \right)^2 }{ \left(n_{\Bar{B}}\mathbb{V}_{\infty} \right)^2 \left( 1 + n_{\Bar{B}} n_{B}^3 \Upsilon(d)^2 n d^\kappa  \right)} \right),
    \end{align*}
    \label{infConcentration}
    with $K_\rho(\infty) = \frac{1}{\ln(\rho^{-1})}, \, \tilde{d} = K_\rho(\infty) \ln(n) \; \text{and} \; d = \left\lceil \tilde{d} \right\rceil = \left\lceil \frac{\ln(n)}{\ln(\rho^{-1})} \right\rceil$.
\end{theorem}

\begin{proof}
We use the previous Lemma \ref{concentrationSI} and set $\tilde{d}=\frac{\ln(n)}{\ln(\rho^{-1})}$ and $d=\left\lceil \tilde{d} \right\rceil$.
\\
Therefore
\begin{align*}
   & d^\kappa =  \left\lceil \frac{\ln(n)}{\ln(\rho^{-1})} \right\rceil ^\kappa. \\
   & \rho^d \leq \rho^{\tilde{d}} = \exp(- \ln(\rho^{-1}) d) = \frac{1}{n}.
\end{align*}
From Lemma \ref{notation}, we get $ N_2 \leq N_1 n_d \leq n_{\Bar{B}} n_{B} d^\kappa n$. Then, according to Lemma \ref{concentrationSI}, it holds
\begin{align*}
& \forall \varepsilon > 2 n n_{\Bar{B}} \rho^{d} \mathbb{V}_{\infty}, \pr \left( |\SI - \esp \left[ \SI \right]  | \geq \varepsilon \right) \leq 2 \exp \left( \frac{-2\left(\varepsilon - 2n n_{\Bar{B}} \rho^{d} \mathbb{V}_{\infty} \right)^2 }{ \left(n_{\Bar{B}}\mathbb{V}_{\infty} \right)^2 \left( n^2 \rho^{2d} + N_2 \left( n_{B} \Upsilon(d) \right)^2 \right)} \right) \\
 & \Rightarrow \forall \varepsilon > 2 n_{\Bar{B}} \mathbb{V}_{\infty}, \, \pr \left( |\SI - \esp \left[ \SI \right]  | \geq \varepsilon \right) \leq 2 \exp \left( \frac{-2\left(\varepsilon - 2  n_{\Bar{B}} \mathbb{V}_{\infty} \right)^2 }{ \left(n_{\Bar{B}}\mathbb{V}_{\infty} \right)^2 \left( 1 + n_{\Bar{B}} n_{B}^3 \Upsilon(d)^2 n d^\kappa  \right)} \right).
\end{align*}

\end{proof}

\begin{rem}
The notation $K_\rho(\infty) = \frac{1}{\ln(\rho^{-1})}$ has been introduced in anticipation to Corollary \ref{mDeviation} and the notation $K_\rho(m) = \frac{1 - \frac{1}{m}}{\ln(\rho^{-1})}$.
\end{rem}

\begin{theorem}[Improved concentration inequality for $\SI$, weak contraction case]
    If we assume $(\bf{H_1^m})$,  $(\bf{H_2^m})$ $(\bf{H_3})$ and $(\bf{H_4})$. It holds, for $\varepsilon \geq 4 n_{\Bar{B}} n_{B} d^{2\kappa} \rho^{-1} n^{\frac{2}{m}} L(n)$
\begin{align*}
     & \pr \left( |\SI - \esp \left[ \SI \right]  | \geq \varepsilon \right) \nonumber \\
     & \leq 2 \exp \left(  \frac{- 2 \rho^2 \left( \frac{\varepsilon}{2} - 2 n_{\Bar{B}} n_{B} d^{2\kappa} \rho^{-1} n^{\frac{2}{m}} L(n) \right)^2}{\left( n_{\Bar{B}} \mathbb{V}_{m} n^{\frac{2}{m}} \right)^2 \left( 1 + n n_{\Bar{B}} n_{B}^3 \Upsilon(d)^2 d^\kappa \right)}\right) + \frac{\rho^m}{n} \left( 2 n_{\Bar{B}} n_{B} d^\kappa + \left( \frac{\mathbb{V}_{m}}{2 n_B d^{2 \kappa} L(n)} \right)^m \right).
\end{align*}
With $K_\rho(m) = \frac{1 - \frac{1}{m}}{ \ln(\rho^{-1})}, \, \tilde{d} = K_\rho(m) \ln(n), \, d =  \left\lceil \tilde{d} \right\rceil = \left\lceil \frac{\left (1 - \frac{1}{m} \right) \ln(n)}{ \ln(\rho^{-1})} \right\rceil$.
\\
And $L(n) = \left( \frac{n_{\Bar{B}} \mathbb{V}_{m}}{\rho} \right) \left( \frac{1}{d^\kappa ( n_{\Bar{B}} n_{B} d^\kappa)^{\frac{1}{m}} n} + n_{\Bar{B}} n_{B}^2 \Upsilon(d)  \right) +\frac{2M}{d^\kappa n^{\frac{2}{m}}}$.
\label{mConcentration}
\end{theorem}

The proof can be found in the Appendix \ref{proofMConcentration}. 

\begin{rem}
In Theorems \ref{infConcentration} and \ref{mConcentration}, quantities $n_{\Bar{B}}, {n_\mathcal{B}},\mathbb{V}_{m}, \mathbb{V}_{\infty}, \kappa$ and $\rho$ are constants and the function $\Upsilon(d)$ can always be bound by a constant $\upsilon$ independent from $d$ (see Lemma \ref{lem:technical}).
\end{rem}

\begin{rem}
In Theorem \ref{mConcentration}, an extra additive term appears. this term decreases quickly with $n$ and thus is often not harmful in practice.
\\
Moreover, adding suitable hypotheses on m can yield to a fully exponential concentration inequality. 
\end{rem}

\subsection{Expected deviation bounds}

Expected deviation bounds may be obtained from previous theorems. In this case, we want to bound $\esp[|\SI - \esp \left[ \SI \right]|]$. 

\begin{cor}
We assume $(\bf{H_1^\infty})$, $(\bf{H_2^\infty})$ and $(\bf{H_3})$, it holds
\begin{equation*}
    \esp[|\SI - \esp \left[ \SI \right]|] \leq n_{\Bar{B}} \mathbb{V}_{\infty} \left( 2 + \sqrt{\frac{\pi}{2} \left( 1  + n_{\Bar{B}} n_{B}^3 \Upsilon(d)^2 d^\kappa n \right)} \right),
\end{equation*}
 with $K_\rho(\infty) = \frac{1}{\ln(\rho^{-1})}$ and $d = \left\lceil K_\rho(\infty) \ln(n) \right\rceil$.
\label{infDeviation}
\end{cor}

\begin{proof}

\begin{align*}
    \esp[|\SI - \esp \left[ \SI \right]|] &= \int_{0}^{\infty} \pr \left( |\SI - \esp \left[ \SI \right]  | \geq t \right) dt \\
    &= \int_0^{2 n_{\Bar{B}} \mathbb{V}_{\infty}} \pr \left( |\SI - \esp \left[ \SI \right]  | \geq t \right) dt + \int_{2 n_{\Bar{B}} \mathbb{V}_{\infty}}^{\infty} \pr \left( |\SI - \esp \left[ \SI \right]  | \geq t \right) dt \\
    & \leq 2 n_{\Bar{B}} \mathbb{V}_{\infty} + \int_{2 n_{\Bar{B}} \mathbb{V}_{\infty}}^{\infty} 2 \exp \left( \frac{-2\left(t - 2 n_{\Bar{B}} \mathbb{V}_{\infty} \right)^2 }{ \left( n_{\Bar{B}} \mathbb{V}_{\infty} \right)^2 \left( 1  + n_{\Bar{B}} n_{B}^3 \Upsilon(d)^2 d^\kappa n \right)} \right) dt \\
    & \leq 2 n_{\Bar{B}} \mathbb{V}_{\infty} + n_{\Bar{B}} \mathbb{V}_{\infty} \sqrt{\frac{\pi}{2} \left( 1  + n_{\Bar{B}} n_{B}^3 \Upsilon(d)^2 d^\kappa n \right)}.
\end{align*}
\end{proof}

\begin{cor}
We assume $(\bf{H_1^m})$,  $(\bf{H_2^m})$, $(\bf{H_3})$ and $(\bf{H_4})$, it holds
\begin{align*}
    \esp[|\SI - \esp \left[ \SI \right]|]
    \leq & 2 \frac{n_{\Bar{B}} \mathbb{V}_{m} n^{\frac{2}{m}}}{\rho} \sqrt{\frac{\pi}{2} \left( 1 + n n_{\Bar{B}} n_{B}^3 \Upsilon(d)^2 d^\kappa \right)} + 4 n_{\Bar{B}} n_{B} d^{2\kappa} \rho^{-1} n^{\frac{2}{m}} L(n) \\ 
    & + 2\rho^m M \left( 2 n_{\Bar{B}} n_{B} d^\kappa + \left( \frac{\mathbb{V}_{m}}{2 n_B d^{2 \kappa} L(n)} \right)^m \right),
\end{align*}
with $K_\rho(m) = \frac{1 - \frac{1}{m}}{\ln(\rho^{-1})}, d = \left\lceil K_\rho(m) \ln(n) \right\rceil $ and $L(n) =  \left( \frac{n_{\Bar{B}} \mathbb{V}_{m}}{\rho} \right) \left( \frac{1}{d^\kappa ( n_{\Bar{B}} n_{B} d^\kappa)^{\frac{1}{m}} n} + n_{\Bar{B}} n_{B}^2 \Upsilon(d)  \right) +\frac{2M}{d^\kappa n^{\frac{2}{m}}}$.
\label{mDeviation}
\end{cor}

\begin{proof}
\begin{align*}
    & \esp[|\SI - \esp \left[ \SI \right]|] \\
    &= \int_{0}^{\infty} \pr \left( |\SI - \esp \left[ \SI \right]  | \geq t \right) dt \\
    &=  \int_{0}^{4 n_{\Bar{B}} n_{B} d^{2\kappa} \rho^{-1} n^{\frac{2}{m}} L(n)} \pr \left( |\SI - \esp \left[ \SI \right]  | \geq t \right) dt + \int_{4 n_{\Bar{B}} n_{B} d^{2\kappa} \rho^{-1} n^{\frac{2}{m}} L(n)}^{\infty} \pr \left( |\SI - \esp \left[ \SI \right]  | \geq t \right)dt \\
    & \leq 4 n_{\Bar{B}} n_{B} d^{2\kappa} \rho^{-1} n^{\frac{2}{m}} L(n) + \int_{4 n_{\Bar{B}} n_{B} d^{2\kappa} \rho^{-1} n^{\frac{2}{m}} L(n)}^{2nM} \pr \left( |\SI - \esp \left[ \SI \right]  | \geq t \right)dt \\ 
    & \text{(because $\SI \leq nM$ (see hypothesis $(\bf{H_4})$))} \\
    & \leq 4 n_{\Bar{B}} n_{B} d^{2\kappa} \rho^{-1} n^{\frac{2}{m}} L(n) + 
    \int_{4 n_{\Bar{B}} n_{B} d^{2\kappa} \rho^{-1} n^{\frac{2}{m}} L(n)}^{2nM}  2 \exp \left(  \frac{- 2 \rho^2 \left( \frac{t}{2} - 2 n_{\Bar{B}} n_{B} d^{2\kappa} \rho^{-1} n^{\frac{2}{m}} L(n) \right)^2}{\left( n_{\Bar{B}} \mathbb{V}_{m} n^{\frac{2}{m}} \right)^2 \left( 1 + n n_{\Bar{B}} n_{B}^3 \Upsilon(d)^2 d^\kappa \right)}\right) dt \\
    & + 2 M \rho^m \left( 2 n_{\Bar{B}} n_{B} d^\kappa + \left( \frac{\mathbb{V}_{m}}{2 n_B d^{2 \kappa} L(n)} \right)^m \right).
\end{align*}
Moreover,
\begin{align*}
    & \int_{4 n_{\Bar{B}} n_{B} d^{2\kappa} \rho^{-1} n^{\frac{2}{m}} L(n)}^{2nM}  2 \exp \left(  \frac{- 2 \rho^2 \left( \frac{t}{2} - 2 n_{\Bar{B}} n_{B} d^{2\kappa} \rho^{-1} n^{\frac{2}{m}} L(n) \right)^2}{\left( n_{\Bar{B}} \mathbb{V}_{m} n^{\frac{2}{m}} \right)^2 \left( 1 + n n_{\Bar{B}} n_{B}^3 \Upsilon(d)^2 d^\kappa \right)}\right) dt \\
    & = 4 \int_{0}^{2nM} \exp \left(  \frac{- 2 \rho^2 t^2}{\left( n_{\Bar{B}} \mathbb{V}_{m} n^{\frac{2}{m}} \right)^2 \left( 1 + n n_{\Bar{B}} n_{B}^3 \Upsilon(d)^2 d^\kappa \right)}\right) dt
    \leq 2 \frac{n_{\Bar{B}} \mathbb{V}_{m} n^{\frac{2}{m}}}{\rho} \sqrt{\frac{\pi}{2} \left( 1 + n n_{\Bar{B}} n_{B}^3 \Upsilon(d)^2 d^\kappa \right)}.
\end{align*}
Therefore, 
\begin{align*}
    \esp[|\SI - \esp \left[ \SI \right]|] 
    \leq & 4 n_{\Bar{B}} n_{B} d^{2\kappa} \rho^{-1} n^{\frac{2}{m}} L(n) +  2 \frac{n_{\Bar{B}} \mathbb{V}_{m} n^{\frac{2}{m}}}{\rho} \sqrt{\frac{\pi}{2} \left( 1 + n n_{\Bar{B}} n_{B}^3 \Upsilon(d)^2 d^\kappa \right)} \\
    & + 2\rho^m M \left( 2 n_{\Bar{B}} n_{B} d^\kappa + \left( \frac{\mathbb{V}_{m}}{2 n_B d^{2 \kappa} L(n)} \right)^m \right).
\end{align*}

\end{proof}

Using these moment inequalities, we can also analyze the limit behavior of  $\esp[|\SI - \esp \left[ \SI \right]|]$.

\begin{cor}
~\begin{itemize}
    \item We assume $(\bf{H_1^\infty})$, $(\bf{H_2^\infty})$ and $(\bf{H_3})$. It holds
    \begin{align*}
        \esp[|\SI - \esp \left[ \SI \right]|] \leq G_1(\kappa, \rho, \mathbb{V}_{\infty}, n_{\Bar{B}}, n_{B}, n) & \underset{n \to \infty}{\sim} 
        n_{B} n_{\Bar{B}} \mathbb{V}_{\infty} \upsilon \sqrt{\frac{\pi}{2} n_{B} n_{\Bar{B}} (K_\rho(\infty))^\kappa {n \ln(n)}^\kappa}, \\ & \underset{n \to \infty}{=} \mathcal{O}\left( \sqrt{n {\ln(n)}^\kappa } \right),
    \end{align*}
    with $K_\rho(\infty) = \frac{1}{\ln(\rho^{-1})}$.
    \item We assume $(\bf{H_1^m})$, $(\bf{H_2^m})$, $(\bf{H_2})$ and $(\bf{H_3})$. It holds
    \begin{align*}
        \esp[|\SI - \esp \left[ \SI \right]|] \leq G_2(\kappa, \rho, \mathbb{V}_{m}, n_{\Bar{B}}, n_{B}, n) 
        & \underset{n \to \infty}{\sim}
        \frac{n_{\Bar{B}} n_B \mathbb{V}_{m} \upsilon n^{\frac{2}{m}}}{\rho} \sqrt{2 \pi n_{\Bar{B}} n_B K_\rho(m)^\kappa \ln(n)^\kappa n}
        \\
        & \underset{n \to \infty}{=} \mathcal{O} \left( n^{\frac{2}{m}} \sqrt{n {\ln(n)}^\kappa} \right),
    \end{align*}
    with $K_\rho(m) = \frac{1 - \frac{1}{m}}{\ln(\rho^{-1})}$.
\end{itemize}
\label{asymptoticAnalys}
\end{cor}
\begin{proof}
We use Corollaries \ref{infDeviation} and \ref{mDeviation} and the bound $\forall d \in \mathbb{N}, \Upsilon(d) \leq \upsilon$ (see Lemma \ref{lem:technical}).
\end{proof}

\begin{rem}
Theorem \ref{mConcentration}, Corollaries \ref{mDeviation} and \ref{asymptoticAnalys} with hypotheses $(\bf{H_1^m})$ and $(\bf{H_2^m})$ are useful only if $m>4$. If it is not the case $\displaystyle \lim_{n \to +\infty}n^{\frac{2}{m} -\frac{1}{2}} \ne 0$, consequently $\displaystyle \lim_{n \to +\infty} \esp\left[\frac{|\SI - \esp \left[ \SI \right]|}{n}\right] \ne 0 $. In the context of learning theory, it means that the empirical risk may not converge to theoretical risk (see Section \ref{subsec:Learning}).
\end{rem}

\newpage

\appendix
\section{Proof of Corollary \ref{expMoment}}
\label{proofExpMoment}
\begin{proof}
    ~\paragraph{If $(\bf{H_1^\infty})$, $(\bf{H_2^\infty})$ and $(\bf{H_3})$ are verified,}
    \begin{align*}
        & \esp[\exp(s | \rth{\model} - \remp{\model} |)] = \int_{t=0}^\infty \pr \left( \exp(s | \rth{\model} - \remp{\model} |) \geq t \right) dt \\
        & = \int_{t=0}^{\exp \left( \frac{2 s n_{\Bar{B}} \mathbb{V}_{\infty} }{n} \right)} \pr \left( \exp(s | \rth{\model} - \remp{\model} |) \geq t \right) + \int_{t=\exp \left( \frac{2 s n_{\Bar{B}} \mathbb{V}_{\infty} }{n} \right)}^{\infty} \pr \left( \exp(s | \rth{\model} - \remp{\model} |) \geq t \right) dt \\
        & = \exp \left( \frac{2 s n_{\Bar{B}} \mathbb{V}_{\infty} }{n} \right) + \int_{t=\exp \left( \frac{2 s n_{\Bar{B}} \mathbb{V}_{\infty} }{n} \right)}^{\infty} \pr \left( | \rth{\model} - \remp{\model} | \geq \frac{\ln(t)}{s} \right) dt \\
        & = \exp \left( \frac{2 s n_{\Bar{B}} \mathbb{V}_{\infty} }{n} \right) + \int_{t=\exp \left( \frac{2 s n_{\Bar{B}} \mathbb{V}_{\infty} }{n} \right)}^{\infty} \exp \left(\frac{ -2 n^2 \left(  \frac{\ln(t)}{s} - \frac{2  n_{\Bar{B}} \mathbb{V}_{\infty}}{n}  \right)^2  }{\left( n_{\Bar{B}} \mathbb{V}_{\infty} \right)^2 \left( 1 + A n_{\Bar{B}} n_B^3 \kappa!^2 \left\lceil \ln(n) \right\rceil^\kappa n \right)} \right) dt \quad \text{using Theorem \ref{ErrGenTail}}\\
        & \leq \exp \left( \frac{2 s n_{\Bar{B}} \mathbb{V}_{\infty} }{n} \right) \left( 1 + \int_{t=1}^{\infty} 2 \exp \left(\frac{ -2 n^2  \ln(t)^2  }{ \left( n_{\Bar{B}} \mathbb{V}_{\infty} s \right)^2 \left( 1 + A n_{\Bar{B}} n_B^3 \kappa!^2 \left\lceil \ln(n) \right\rceil^\kappa n \right) } \right) dt \right).
    \end{align*}
    Then, 
    \begin{align*}
        & \int_{t=1}^{\infty} 2 \exp \left(\frac{ -2 n^2  \ln(t)^2  }{ \left( n_{\Bar{B}} \mathbb{V}_{\infty} s \right)^2 \left( 1 + A n_{\Bar{B}} n_B^3 \kappa!^2 \left\lceil \ln(n) \right\rceil^\kappa n \right) } \right) dt \\
        & = 2 \int_{t=0}^{\infty} \exp \left(\frac{ -2 n^2  t^2  }{ \left( n_{\Bar{B}} \mathbb{V}_{\infty} s \right)^2 \left( 1 + A n_{\Bar{B}} n_B^3 \kappa!^2 \left\lceil \ln(n) \right\rceil^\kappa n \right) } + t \right) dt \\
        & \leq \frac{n_{\Bar{B}} \mathbb{V}_{\infty} s}{n} \sqrt{2 \pi \left( 1 + A n_{\Bar{B}} n_B^3 \kappa!^2 \left\lceil \ln(n) \right\rceil^\kappa n \right)} \exp \left( \frac{\left( n_{\Bar{B}} \mathbb{V}_{\infty} s \right)^2 \left( 1 + A n_{\Bar{B}} n_B^3 \kappa!^2 \left\lceil \ln(n) \right\rceil^\kappa n \right)}{8 n^2} \right).
    \end{align*}
    It yields,
    \begin{align*}
        & \esp[\exp(s | \rth{\model} - \remp{\model} |)] \\ & \leq \exp \left( \frac{2 s n_{\Bar{B}} \mathbb{V}_{\infty} }{n} + \frac{\left( n_{\Bar{B}} \mathbb{V}_{\infty} s \right)^2 \left( 1 + A n_{\Bar{B}} n_B^3 \kappa!^2 \left\lceil \ln(n) \right\rceil^\kappa n \right)}{8 n^2} \right) \left( 1 + \frac{n_{\Bar{B}} \mathbb{V}_{\infty} s}{n} \sqrt{2 \pi \left( 1 + A n_{\Bar{B}} n_B^3 \kappa!^2 \left\lceil \ln(n) \right\rceil^\kappa n \right)}  \right)
    \end{align*}
    \paragraph{If $(\bf{H_1^m})$, $(\bf{H_2^m})$, $(\bf{H_3})$ and $(\bf{H_4})$ are verified.}
    The demonstration is very similar to the previous point. Indeed, we have,
    \begin{align*}
        & \esp[\exp(s | \rth{\model} - \remp{\model} |)] \\ 
        & \leq \exp \left( \frac{2sL_1(n)}{n^{1-\frac{2}{m}}} \right) + \int_{t=\exp \left( \frac{2sL_1(n)}{n^{1-\frac{2}{m}}} \right)}^{\exp(Ms)} \pr \left( | \rth{\model} - \remp{\model} | \geq \frac{\ln(t)}{s} \right) dt \\
        & \leq \exp \left( \frac{2sL_1(n)}{n^{1-\frac{2}{m}}} \right) + \int_{t=\exp \left( \frac{2sL_1(n)}{n^{1-\frac{2}{m}}} \right)}^{\exp(Ms)}  2 \exp \left(  \frac{-  2 n^{2-\frac{4}{m}} \left( \frac{\ln(t)}{2s} - \frac{L_1(n)}{n^{1-\frac{2}{m}}}  \right)^2}{ \left(H n_{\Bar{B}} \right)^2 \left(1 + E n_{\Bar{B}} n_B^3 (\kappa!)^2 \left\lceil \ln(n) \right\rceil^\kappa n \right) }\right) + \frac{\rho^m L_2(n)}{n}dt
    \end{align*}
    And, 
    \begin{align*}
        & \int_{t=\exp \left( \frac{2sL_1(n)}{n^{1-\frac{2}{m}}} \right)}^{\exp(Ms)}  2 \exp \left(  \frac{-  2 n^{2-\frac{4}{m}} \left( \frac{\ln(t)}{2s} - \frac{L_1(n)}{n^{1-\frac{2}{m}}}  \right)^2}{ \left(H n_{\Bar{B}} \right)^2 \left(1 + E n_{\Bar{B}} n_B^3 (\kappa!)^2 \left\lceil \ln(n) \right\rceil^\kappa n \right) }\right) dt \\
        & = \exp \left( \frac{2sL_1(n)}{n^{1-\frac{2}{m}}} \right) \int_{t=1}^{\infty} 2 \exp \left(  \frac{- n^{2-\frac{4}{m}} \ln(t)^2 }{ 2 \left(H n_{\Bar{B}} s \right)^2 \left(1 + E n_{\Bar{B}} n_B^3 (\kappa!)^2 \left\lceil \ln(n) \right\rceil^\kappa n \right) }\right) dt \\
        & = 2 \exp \left( \frac{2sL_1(n)}{n^{1-\frac{2}{m}}} \right) \int_{t=0}^{\infty} \exp \left(  \frac{- n^{2-\frac{4}{m}} t^2 }{ 2 \left(H n_{\Bar{B}} s \right)^2 \left(1 + E n_{\Bar{B}} n_B^3 (\kappa!)^2 \left\lceil \ln(n) \right\rceil^\kappa n \right) } + t \right) dt \\
        & = 2 \exp \left( \frac{2sL_1(n)}{n^{1-\frac{2}{m}}} \right) \frac{H n_{\Bar{B}} s}{n^{1-\frac{2}{m}}} \sqrt{2 \pi \left(1 + E n_{\Bar{B}} n_B^3 (\kappa!)^2 \left\lceil \ln(n) \right\rceil^\kappa n \right)} \exp \left( \frac{\left(H n_{\Bar{B}} s \right)^2 \left(1 + E n_{\Bar{B}} n_B^3 (\kappa!)^2 \left\lceil \ln(n) \right\rceil^\kappa n \right)}{2 n^{2-\frac{4}{m}}} \right) \\
        & \leq 2 \exp \left( \frac{1}{2 n^{1-\frac{4}{m}}} \left( 4sL_1(n) + \left(H n_{\Bar{B}} s \right)^2 \left(1 + E n_{\Bar{B}} n_B^3 (\kappa!)^2 \left\lceil \ln(n) \right\rceil^\kappa n \right) \right) \right) \\
        & \times \frac{H n_{\Bar{B}} s}{n^{1-\frac{2}{m}}} \sqrt{2 \pi \left(1 + E n_{\Bar{B}} n_B^3 (\kappa!)^2 \left\lceil \ln(n) \right\rceil^\kappa n \right)}
    \end{align*}
    Consequently,
    \begin{align*}
        & \esp[\exp(s | \rth{\model} - \remp{\model} |)] 
        \leq
        \exp \left( \frac{1}{2 n^{1-\frac{4}{m}}} \left( 4sL(n) + \left(H n_{\Bar{B}} s \right)^2 \left(1 + E n_{\Bar{B}} n_B^3 (\kappa!)^2 \left\lceil \ln(n) \right\rceil^\kappa n \right) \right) \right) \\
        &\times \left( 1 + \frac{\exp \left( Ms \right) \rho^m L_2(n)}{n}  
        + \frac{2H n_{\Bar{B}} s}{n^{1-\frac{2}{m}}} \sqrt{2 \pi \left(1 + E n_{\Bar{B}} n_B^3 (\kappa!)^2 \left\lceil \ln(n) \right\rceil^\kappa n \right)}
        \right).
    \end{align*}
\end{proof}
\section{Proof of Lemma \ref{lem:approxF}}
\begin{proof}
\label{proofApprox}
Let's prove by induction that for each $i$ in $[0, d]$, 
\begin{equation}
     \forall t \in \mathbb{Z}^\kappa, \forall d \in \mathbb{N}, \|X_t - \Xt{t}{i} \|_m \leq \rho^{i} \mathbb{V}_m.
    \label{recu}
\end{equation}
For $i=0$, $\Xt{t}{0} = \Bar{X}$ and $\Bar{X}$ is drawn from the law $\mu_X$. Therefore, using hypothesis $(\bf{H_2^m})$, we get 
\begin{equation*}
    \forall t \in \mathbb{Z}^\kappa, \|X_t - \Xt{t}{0} \|_m \leq \mathbb{V}_m.
\end{equation*}
Thus Equation \eqref{recu} is verified for $i=0$.
\\
Moreover, if we suppose that \eqref{recu} is verified for $i$. For each $t \in \mathbb{Z}^\kappa$, it holds
\begin{align*}
    \|X_t - \Xt{t}{i+1} \|_m &\leq \| F \left( (X_{t+s})_{s \in \mathcal{B}}, \varepsilon_t \right) -  F \left( (\Xt{t+s}{i})_{s \in \mathcal{B}}, \varepsilon_t \right)\|_m \\
    &\leq \sum_{s \in \mathcal{B}} \lambda_s \|X_{t+s} - \Xt{t+s}{i}\|_m \leq \sum_{s \in \mathcal{B}} \lambda_s \rho^i \mathbb{V}_m \leq \rho^{i+1} \mathbb{V}_m.
\end{align*}
Consequently, by induction, Equation \eqref{recu} is verified for each $i$ in $[0,d]$. Finally, setting $i=d$ yields Lemma 1.
\end{proof}
\section{Proof of Lemma \ref{momentbdi}}
\begin{proof}
\label{proveBDI}
We prove the result for $(\bf{H_1^m})$ and $(\bf{H_2^m})$, the case $(\bf{H_1^\infty})$ and $(\bf{H_2^\infty})$ is similar.

~\paragraph{We first show that $\| \SItilde - \SItildep \|_m \leq n n_{\Bar{B}} \rho^d {\mathbb{V}_m}$.}

\begin{align*}
\| \SItilde - \SItildep \|_m & =  \| \sum_{t \in \mathcal{I}} \Phi((\Xt{t+s}{d})_{s \in \Bar{\mathcal{B}}}) - \sum_{t \in \mathcal{I}} \Phi((\Xtt{t+s}{d})_{s \in \Bar{\mathcal{B}}}) \|_m \\
& \leq \sum\limits_{t \in \mathcal{\mathcal{I}}} \| \Phi((\Xt{t+s}{d})_{s \in \Bar{\mathcal{B}}}) - \Phi((\Xtt{t+s}{d})_{s \in \Bar{\mathcal{B}}})  \|_m \\
&\leq \sum\limits_{t \in \mathcal{\mathcal{I}}} \sum\limits_{s \in \Bar{\mathcal{B}}} \|\Xt{t+s}{d} - \Xtt{t+s}{d} \|_m \\
& \leq n n_{\Bar{B}} \rho^d {\mathbb{V}_m}.
\end{align*}

\paragraph{We then bound $\| \SItilde - \SItildei \|_m$.}
For all $t$ and $i$ $\mathbb{Z}^\kappa$ we have the following properties
\[
\Xt{t}{d} \ne \Xt{t}{d,i} \iff i \in \mathcal{V}(d\delta,t) \iff t \in \mathcal{V}(d\delta,i).
\] 

\begin{align*}
    \| \SItildei - \SItilde \|_m & =   \| \sum_{t \in \mathcal{I}} \Phi((\Xt{t+s}{d})_{s \in \Bar{\mathcal{B}}}) - \sum_{t \in \mathcal{I}} \Phi((\Xt{t+s}{d,i})_{s \in \Bar{\mathcal{B}}}) \|_m
    \\
    & \leq \sum\limits_{t \in \mathcal{\mathcal{I}}} \sum\limits_{s \in \Bar{\mathcal{B}}} \|\Xt{t+s}{d} - \Xt{t+s}{d,i} \|_m   
    \\
    & \leq \sum\limits_{t \in \mathbb{Z}^\kappa} \sum\limits_{s \in \Bar{\mathcal{B}} + t} \|\Xt{s}{d} - \Xt{s}{d,i} \|_m .
\end{align*}
Only random variables $\Xt{t}{d,i}$ with $t \in \mathcal{V}(d\delta,i)$ are impacted by the replacement of $\varepsilon_i$ by $\varepsilon_i^\prime$. Thus, at worst, each $t \in \mathcal{V}(d\delta,i)$ appears $n_{\Bar{B}}$ times in the sum. Therefore, we get
\begin{align*}
    \| \SItildei - \SItilde \|_m & \leq  \sum_{t \in \mathcal{V}(d\delta, i)} n_{\Bar{B}}\| \Xt{t}{d,i} - \Xt{t}{d} \|_m.
\end{align*}
Moreover, for all $t$ in $\mathbb{Z}^\kappa$
\begin{equation*}
    \mathcal{V}(d\delta, t) = \bigcup_{c=0}^d \mathcal{V} \left( c\delta, t \right) \backslash \mathcal{V} \left( (c-1) \delta, t \right) \quad  \text{with} \quad \mathcal{V}(0, t) = {t} \quad \text{and} \quad \mathcal{V}(-1, t) =  \emptyset.
\end{equation*}
And by definition of $\mathcal{V}(d\delta, t)$ (see Section \ref{1notations}), it holds
\begin{equation*}
\forall r_1 \ne r_2, \left( \mathcal{V}(r_1\delta, t) \backslash \mathcal{V}((r_1-1)\delta, t) \right) \bigcap \left( \mathcal{V}(r_2\delta, t) \backslash \mathcal{V}((r_2-1)\delta, ts) \right) = \emptyset.
\end{equation*}
Therefore, $\displaystyle \bigcup_{c=0}^d \mathcal{V}(c\delta, t) \backslash \mathcal{V}(c-1, t)$ is a partition of the set $\mathcal{V}(d\delta, t)$ and we can rewrite the previous inequality as
\begin{align*}
    \| \SItildei - \SItilde \|_m & \leq n_{\Bar{B}}\sum_{c=0}^d \left( \sum_{s \in \mathcal{V}(c\delta, i) \backslash \mathcal{V}((c-1)\delta, i)} \| \Xt{s}{d,i} - \Xt{s}{d}\|_m \right).
\end{align*}
Using Lemma \ref{bdiXt}. We get
\begin{equation*}
     \| \SItildei - \SItilde \|_m \leq n_{\Bar{B}}\sum_{c=0}^d \card{ \mathcal{V}(c\delta, i) \backslash \mathcal{V}((c-1)\delta, i) } \rho^c \mathbb{V}_m.
\end{equation*}
$\forall c \in \mathbb{N}, \mathcal{V}(c \delta,i)$ is a $\kappa-$orthotope and
\begin{equation*}
    \card{\mathcal{V}(c \delta,i)} = \prod_{j=1}^\kappa \left( 2c\delta_j + 1 \right).
\end{equation*}
Consequently, for all $c>1$,
\begin{align*}
    \card{\mathcal{V}(c\delta, i) \backslash \mathcal{V}((c-1)\delta, i)} &= \card{\mathcal{V}(c\delta,i)} - \card{\mathcal{V}((c-1)\delta,i)} \\
    &= \prod_{j=1}^\kappa \left( 2c\delta_j + 1 \right) - \prod_{j=1}^\kappa \left( 2(c-1)\delta_j + 1 \right) \\
    & \leq c^\kappa  \prod_{j=1}^\kappa \left( 2\delta_j + \frac{1}{c} \right) - (c-1)^\kappa\prod_{j=1}^\kappa \left( 2\delta_j + \frac{1}{c-1} \right) \\
    & \leq \prod_{j=1}^\kappa \left( 2\delta_j + \frac{1}{c-1} \right) \left( c^\kappa - (c-1)^\kappa \right) \\
    & \leq \prod_{j=1}^\kappa \left( 2\delta_j + 1 \right) \left( c^\kappa - (c-1)^\kappa \right) \\
    & \leq n_{\mathcal{B}}\left( c^\kappa - (c-1)^\kappa \right) \leq n_{\mathcal{B}} \kappa c^{\kappa - 1} \text{ with $n_{\mathcal{B}} = \prod_{j=1}^\kappa \left( 2\delta_j + 1 \right) = \card{\mathcal{B}} + 1$}.
\end{align*}
Eventually, we get:
$ \displaystyle
    \| \SItildei - \SItilde \|_m \leq n_{\Bar{B}} \mathbb{V}_m \left(1 +\sum_{c=1}^d n_{\mathcal{B}} \kappa c^{\kappa - 1} \rho^c \right).$
\end{proof}
\section{Proof of Lemma \ref{lem:technical}}
\begin{proof}
\label{techniLemma}
Let $p \in \mathbb{N}$ and $(a,b) \in\mathbb{R}^2$. \\
We first compute $\displaystyle I_p = \int_a^b t^p \rho^t dt$.
\begin{equation*}
    I_p = \int_a^b t^p \rho^t dt = \left[ t^p \frac{\rho^t}{\ln(\rho)} \right]_a^b + \frac{p}{\ln(\rho^{-1})} \int_a^b t^{p-1}  \rho^t dt = \frac{a^p \rho^a - b^p \rho^b}{\ln(\rho^{-1})} + \frac{p}{\ln(\rho^-1)} I_{p-1}.
\end{equation*}
And
\begin{equation*}
    I_0 = \int_a^b \rho^t dt = \frac{\rho^a - \rho^b}{\ln(\rho^{-1})}.
\end{equation*}
By induction, we get 
\begin{equation}
   I_p = \sum_{i=0}^p \frac{a^i\rho^a - b^i\rho^b}{\ln(\rho^{-1})^{p-i+1}} \times \frac{p!}{i!} = \frac{p!}{\ln(\rho^{-1})^{p+1}} \sum_{i=0}^p \frac{\left( a^i\rho^a - b^i\rho^b \right) \ln(\rho^{-1})^i}{i!}.
   \label{IP}
\end{equation}
Let $f(t) = t^p \rho^t$; we have $f^\prime(t) = t^{p-1} \rho^t \left( p - t \ln(\rho^{-1}) \right)$. Then $f^\prime(t)=0 \iff t = \frac{p}{\ln(\rho^{-1})}$.
Thus f is increasing on $[0, \frac{p}{\ln(\rho^{-1})}]$ and decreasing on $[\frac{p}{\ln(\rho^{-1})}, + \infty]$. \\
Applying this to our case, 
\begin{itemize}
    \item if $d < \lfloor \frac{\kappa -1}{\ln(\rho^{-1})} \rfloor$. Then: $ \displaystyle \sum_{c=1}^{d} c^{\kappa -1}\rho^c \leq \int_0^{d+1} t^{\kappa -1}\rho^t dt $.
    \item if $d = \lfloor \frac{\kappa -1}{\ln(\rho^{-1})} \rfloor$, Then  $ \displaystyle \sum_{c=1}^{d} c^{\kappa -1}\rho^c \leq \int_0^{\lfloor \frac{\kappa -1}{\ln(\rho^{-1})} \rfloor} t^{\kappa -1}\rho^t dt + \left( \frac{\kappa-1}{\ln(\rho{-1})e} \right)^{\kappa-1} $.
    \item if $d > \lfloor \frac{\kappa -1}{\ln(\rho^{-1})} \rfloor$. Then $ \displaystyle \sum_{c=1}^d c^{\kappa -1}\rho^c \leq \int_0^{\lfloor \frac{\kappa -1}{\ln(\rho^{-1})} \rfloor} t^{\kappa -1}\rho^t dt + \left( \frac{\kappa-1}{\ln(\rho{-1})e} \right)^{\kappa-1} + \int_{\lfloor \frac{\kappa -1}{\ln(\rho^{-1})} \rfloor}^{d} t^{\kappa -1}\rho^t dt $.
\end{itemize}
Using Equation \eqref{IP}, we get
\begin{itemize}
    \item $\displaystyle \int_0^{d+1} t^{\kappa -1}\rho^t dt \leq \frac{(\kappa -1)!}{\ln(\rho^{-1})^{\kappa}} \left( 1 - \rho^{d+1} \sum_{i=0}^{\kappa -1} \frac{\left( (d+1) \ln(\rho^{-1}) \right)^i}{i!} \right)$.
    \item $\displaystyle \int_{\left\lfloor \frac{\kappa -1}{\ln(\rho^{-1})} \right\rfloor}^{d} t^{\kappa -1}\rho^t dt \leq \frac{(\kappa -1)!}{\ln(\rho^{-1})^{\kappa}} \left( \rho^{\left\lfloor \frac{\kappa - 1}{\ln(\rho^{-1})} \right\rfloor} \sum_{i=0}^{\kappa -1} \frac{\left( \left\lfloor\frac{\kappa - 1}{\ln(\rho^{-1})} \right\rfloor \ln(\rho^{-1}) \right)^i}{i!} - \rho^d \sum_{i=0}^{\kappa - 1} \frac{\left( d \ln(\rho^{-1}) \right)^i}{i!} \right) $.
\end{itemize}
We recall that if $i \in \mathcal{R}$,
\begin{equation*}
    \| \SItildei - \SItilde \|_m 
     \leq n_{\Bar{B}} \mathbb{V}_m \left(1 + n_{B} \kappa \sum_{c=1}^d c^{\kappa - 1} \rho^c \right) 
     \leq n_{\Bar{B}} n_B \mathbb{V}_m \kappa \left( \frac{1}{n_B \kappa} + \sum_{c=1}^d c^{\kappa - 1} \rho^c \right).
\end{equation*}
We now define the function $\Upsilon$.
\begin{align*}
    \Upsilon: \mathbb{N} &\mapsto \mathbb{R} \\
    \Upsilon(d) &=
    \left\{
    \begin{array}{l}
        \displaystyle
        \kappa \left( \frac{1}{n_B \kappa} + \frac{(\kappa -1)!}{\ln(\rho^{-1})^{\kappa}} \left( 1 - \rho^{d+1} \sum_{i=0}^{\kappa -1} \frac{\left( (d+1) \ln(\rho^{-1}) \right)^i}{i!} \right) \right), \; \text{ if } d < \left\lfloor \frac{\kappa -1}{\ln(\rho^{-1})} \right\rfloor. \\
        \displaystyle
        \kappa \left( \frac{1}{n_B \kappa} + \frac{(\kappa -1)!}{\ln(\rho^{-1})^{\kappa}} \left( 1 - e^{- (\kappa - 1)} \sum_{i=0}^{\kappa -1} \frac{\left( \left\lfloor\frac{\kappa - 1}{\ln(\rho^{-1})} \right\rfloor \ln(\rho^{-1}) \right)^i}{i!} \right) + \left( \frac{\kappa-1}{\ln(\rho^{-1})e} \right)^{\kappa-1} \right), \; \text{ if } d = \left\lfloor \frac{\kappa -1}{\ln(\rho^{-1})} \right\rfloor. \\
        \displaystyle
        \kappa \left( \frac{1}{n_B \kappa} + \frac{(\kappa -1)!}{\ln(\rho^{-1})^{\kappa}} \left( 1 - \rho^d \sum_{i=0}^{\kappa - 1} \frac{\left( d \ln(\rho^{-1}) \right)^i}{i!} \right) + \left( \frac{\kappa-1}{\ln(\rho{-1})e} \right)^{\kappa-1} \right), \; \text{ if } d > \left\lfloor \frac{\kappa -1}{\ln(\rho^{-1})} \right\rfloor.
    \end{array}
    \right.
\end{align*}
We proved that if $i \in \mathcal{R}$,
\begin{equation*}
    \| \SItildei - \SItilde \|_m \leq n_{\Bar{B}} n_B \mathbb{V}_m  \Upsilon(d).
\end{equation*}

It is important to note that we can easily verify that the function $\Upsilon$ is increasing and bounded by its limit when $d \to \infty$. 
It holds
\begin{align*}
    \forall d \in \mathbb{N}, \Upsilon(d) & \leq \lim_{d \to \infty} \Upsilon(d) = \kappa \left( \frac{1}{n_B \kappa} + \frac{(\kappa -1)!}{\ln(\rho^{-1})^{\kappa}} + \left( \frac{\kappa-1}{\ln(\rho{-1})e} \right)^{\kappa-1} \right) \\
    & \leq \frac{1}{n_B} + \frac{\kappa!}{\ln(\rho^{-1})^{\kappa}} + \kappa \left( \frac{\kappa-1}{\ln(\rho{-1})e} \right)^{\kappa-1}.
\end{align*}
Thereafter, we will denote $\upsilon = \frac{1}{n_B} + \frac{\kappa!}{\ln(\rho^{-1})^{\kappa}} + \kappa \left( \frac{\kappa-1}{\ln(\rho{-1})e} \right)^{\kappa-1}$. Moreover, we emphasize that using Stirling formula, we can proved that $\upsilon \underset{\kappa \to \infty}{\sim} \frac{\kappa!}{\ln(\rho^{-1})^{\kappa}}  $.

\end{proof}
\section{Proof of Theorem \ref{mConcentration}}
\label{proofMConcentration}
\begin{proof}
From Lemma \ref{notation}, we get $ N_2 \leq N_1 n_d \leq n_{\Bar{B}} n_{B} d^\kappa n$.
\\
We set
$\displaystyle \tilde{d} = \frac{(1 - \frac{1}{m})\ln(n)}{\ln(\rho^{-1})}, \, d = \left\lceil \tilde{d} \right\rceil, \, t_1 = \frac{n^{1 + \frac{1}{m}} \rho^{\tilde{d}} n_{\Bar{B}} \mathbb{V}_{m}}{\rho ( n_{\Bar{B}} n_{B} d^\kappa)^{\frac{1}{m}}} \; \text{and} \; t_2 = \frac{n^{\frac{2}{m}} n_{B} n_{\Bar{B}} \Upsilon(d) \mathbb{V}_{m}}{\rho}$.
Then
\begin{align*}
    p  & \leq \left( \frac{n n_{\Bar{B}} \rho^{d} \mathbb{V}_{m}}{t_1} \right)^m + N_2 \left( \frac{n_{\Bar{B}} n_{B} \mathbb{V}_m \Upsilon(d) }{t_2} \right)^m \\
    & \leq \left( \frac{n n_{\Bar{B}} \rho^{\tilde{d}} \mathbb{V}_{m}}{t_1} \right)^m + N_2 \left( \frac{n_{\Bar{B}}n_{B} \mathbb{V}_m \Upsilon(d) }{t_2} \right)^m \\
    & \leq \frac{\rho^m n_{\Bar{B}} n_{B} d^\kappa}{n} + n_{\Bar{B}} n_{B} d^\kappa n \frac{\rho^m}{n^2}\\
    & \leq \frac{2 n_{\Bar{B}} n_{B} d^\kappa \rho^m}{n}.
\end{align*}
\\
We also note that $\rho^{\tilde{d}} = \rho^{\tilde{d}} = \exp(-\ln(\rho^{-1}) K_\rho(m) \ln(n)) = \exp(-\left( 1-\frac{1}{m} \right) \ln(n)) = n^{-\left( 1-\frac{1}{m} \right)}$.
\\
On the other hand,
\begin{align*}
    t_1^2 + N_2 t_2^2 & \leq  \left( \frac{n^{1 + \frac{1}{m}} \rho^{\tilde{d}} n_{\Bar{B}} \mathbb{V}_{m}}{\rho ( n_{\Bar{B}} n_{B} d^\kappa)^{\frac{1}{m}}} \right)^2 + n_{\Bar{B}} n_{B} d^\kappa n \left( \frac{n^{\frac{2}{m}} n_{B} n_{\Bar{B}} \Upsilon(d) \mathbb{V}_{m}}{\rho} \right)^2 \\
    & = \left( \frac{n_{\Bar{B}} \mathbb{V}_{m} n^{\frac{2}{m}}}{\rho} \right)^2 \left( \frac{1}{( n_{\Bar{B}} n_{B} d^\kappa)^{\frac{2}{m}}} + n n_{\Bar{B}} n_{B}^3 \Upsilon(d)^2 d^\kappa \right). \\
\end{align*}
And
\begin{align*}
    \Bar{c} = t_1 + N_2 t_2 
    & \leq  \frac{n^{1 + \frac{1}{m}} \rho^{\tilde{d}} n_{\Bar{B}} \mathbb{V}_{m}}{\rho ( n_{\Bar{B}} n_{B} d^\kappa)^{\frac{1}{m}}} + n_{\Bar{B}} n_{B} d^\kappa n \left( \frac{n^{\frac{2}{m}} n_{B} n_{\Bar{B}} \Upsilon(d) \mathbb{V}_{m}}{\rho} \right) \\
    & \leq \left( \frac{n_{\Bar{B}} \mathbb{V}_{m} n^{\frac{2}{m}}}{\rho} \right) \left( \frac{1}{( n_{\Bar{B}} n_{B} d^\kappa)^{\frac{1}{m}}} + n_{\Bar{B}} n_{B}^2 \Upsilon(d) d^\kappa n \right).
\end{align*}
Therefore,
\begin{align*}
    2npM + p\Bar{c} & 
    \leq \frac{2 n_{\Bar{B}} n_{B} d^\kappa \rho^m}{n} \left( \left( \frac{n_{\Bar{B}} \mathbb{V}_{m} n^{\frac{2}{m}}}{\rho} \right) \left( \frac{1}{( n_{\Bar{B}} n_{B} d^\kappa)^{\frac{1}{m}}} + n_{\Bar{B}} n_{B}^2 \Upsilon(d) d^\kappa n \right) +2nM \right) \\
    &\leq 2 n_{\Bar{B}} n_{B} d^\kappa \rho^m \left(\left( \frac{n_{\Bar{B}} \mathbb{V}_{m} n^{\frac{2}{m}}}{\rho} \right) \left( \frac{1}{( n_{\Bar{B}} n_{B} d^\kappa)^{\frac{1}{m}} n} + n_{\Bar{B}} n_{B}^2 \Upsilon(d) d^\kappa \right) +2M \right) \\
    &\leq 2 n_{\Bar{B}} n_{B} d^{2\kappa} \rho^m n^{\frac{2}{m}} \left( \left( \frac{n_{\Bar{B}} \mathbb{V}_{m}}{\rho} \right) \left( \frac{1}{d^\kappa ( n_{\Bar{B}} n_{B} d^\kappa)^{\frac{1}{m}} n} + n_{\Bar{B}} n_{B}^2 \Upsilon(d)  \right) +\frac{2M}{d^\kappa n^{\frac{2}{m}}} \right) \\
    & \leq 2 n_{\Bar{B}} n_{B} d^{2\kappa} \rho^m n^{\frac{2}{m}} L(n).
\end{align*}
With $L(n) = \left( \frac{n_{\Bar{B}} \mathbb{V}_{m}}{\rho} \right) \left( \frac{1}{d^\kappa ( n_{\Bar{B}} n_{B} d^\kappa)^{\frac{1}{m}} n} + n_{\Bar{B}} n_{B}^2 \Upsilon(d)  \right) +\frac{2M}{d^\kappa n^{\frac{2}{m}}} $.
\\
Consequently using $p \leq  \frac{2 n_{\Bar{B}} n_{B} d^\kappa \rho^m}{n}$. It holds $\forall \varepsilon > 4 n_{\Bar{B}} n_{B} d^{2\kappa} \rho^m n^{\frac{2}{m}} L(n)$,
\begin{align*}
     \exp \left( \frac{-2 (\frac{\varepsilon}{2} - (p \Bar{c} + 2npM ) )^2}{t_1^2 + N_2 t_2^2} \right) 
     & \leq 
     \exp \left(  \frac{- 2 \left( \frac{\varepsilon}{2} - 2 n_{\Bar{B}} n_{B} d^{2\kappa} \rho^m n^{\frac{2}{m}} L(n) \right)^2}{\left( \frac{n_{\Bar{B}} \mathbb{V}_{m} n^{\frac{2}{m}}}{\rho} \right)^2 \left( \frac{1}{( n_{\Bar{B}} n_{B} d^\kappa)^{\frac{2}{m}}} + n n_{\Bar{B}} n_{B}^3 \Upsilon(d)^2 d^\kappa \right)}\right).
\end{align*}
In particular; $\forall \varepsilon > 4 n_{\Bar{B}} n_{B} d^{2\kappa} \rho^{-1} n^{\frac{2}{m}} L(n)$,
\begin{equation*}
    \exp \left( \frac{-2 (\frac{\varepsilon}{2} - (p \Bar{c} + 2npM ) )^2}{t_1^2 + N_2 t_2^2} \right)
    \leq 
    \exp \left(  \frac{- 2 \left( \frac{\varepsilon}{2} - 2 n_{\Bar{B}} n_{B} d^{2\kappa} \rho^{-1} n^{\frac{2}{m}} L(n) \right)^2}{\left( \frac{n_{\Bar{B}} \mathbb{V}_{m} n^{\frac{2}{m}}}{\rho} \right)^2 \left( \frac{1}{( n_{\Bar{B}} n_{B} d^\kappa)^{\frac{2}{m}}} + n n_{\Bar{B}} n_{B}^3 \Upsilon(d)^2 d^\kappa \right)}\right).
\end{equation*}
Finally,
\begin{align*}
    \forall \varepsilon >  4 n_{\Bar{B}} n_{B} d^{2\kappa} \rho^{-1} n^{\frac{2}{m}} L(n), \left( \frac{2n n_{\Bar{B}} \rho^{d} \mathbb{V}_{m}}{\varepsilon} \right)^m 
    &\leq \left( \frac{2n n_{\Bar{B}} \rho^{\tilde{d}} \mathbb{V}_{m}}{4 n_{\Bar{B}} n_{B} d^{2\kappa} \rho^{-1} n^{\frac{2}{m}} L(n)} \right)^m \\
    & \leq \frac{\rho^m}{n} \left( \frac{\mathbb{V}_{m}}{2 n_B d^{2 \kappa} L(n)} \right)^m. \\
\end{align*}
Using Equation \eqref{equ:concentrationSI_2}, for $\varepsilon > 4 n_{\Bar{B}} n_{B} d^{2\kappa} \rho^{-1} n^{\frac{2}{m}} L(n)$, we get
\begin{align*}
     & \pr \left( |\SI - \esp \left[ \SI \right]  | \geq \varepsilon \right) \nonumber \\
     & \leq 2 \exp \left(  \frac{- 2 \left( \frac{\varepsilon}{2} - 2 n_{\Bar{B}} n_{B} d^{2\kappa} \rho^{-1} n^{\frac{2}{m}} L(n) \right)^2}{\left( \frac{n_{\Bar{B}} \mathbb{V}_{m} n^{\frac{2}{m}}}{\rho} \right)^2 \left( \frac{1}{( n_{\Bar{B}} n_{B} d^\kappa)^{\frac{2}{m}}} + n n_{\Bar{B}} n_{B}^3 \Upsilon(d)^2 d^\kappa \right)}\right) + \frac{2 n_{\Bar{B}} n_{B} d^\kappa \rho^m}{n} + \frac{\rho^m}{n} \left( \frac{\mathbb{V}_{m}}{2 n_B d^{2 \kappa} L(n)} \right)^m .
\end{align*}
\end{proof}

\bibliographystyle{imsart-number} 
\bibliography{main}

\end{document}